\documentclass[acmsmall]{acmart}
\AtBeginDocument{%
  }

\setcopyright{acmlicensed}
\copyrightyear{2025}
\acmYear{2025}
\acmDOI{XXXXXXX.XXXXXXX}
\acmConference[Conference acronym 'XX]{Make sure to enter the correct
  conference title from your rights confirmation email}{June 03--05,
  2018}{Woodstock, NY}
\acmISBN{978-1-4503-XXXX-X/2025/06}

\usepackage[ruled,vlined]{algorithm2e}
\usepackage{makecell}
\usepackage{subfigure}
\usepackage{longtable}

\begin{document}

\title{Privacy-Utility Trade-off in Data Publication: A Bilevel Optimization Framework with Curvature-Guided Perturbation}

\author{Yi Yin}
\email{Yi.Yin-1@student.uts.edu.au}
\author{Guangquan Zhang}
\email{Guangquan.Zhang@uts.edu.au}
\author{Hua Zuo}
\email{Hua.Zuo@uts.edu.au}
\author{Jie Lu}
\email{Jie.Lu@uts.edu.au}
\affiliation{%
  \institution{University of Technology Sydney}
  \city{Sydney}
  \country{Australia}
}

\renewcommand{\shortauthors}{Yi Yin, Guangquan Zhang, Hua Zuo, and Jie Lu}

\begin{abstract}
Machine learning models require datasets for effective training, but directly sharing raw data poses significant privacy risk such as membership inference attacks (MIA). To mitigate the risk, privacy-preserving techniques such as data perturbation, generalization, and synthetic data generation are commonly utilized. However, these methods often degrade data accuracy, specificity, and diversity, limiting the performance of downstream tasks and thus reducing data utility. Therefore, striking an optimal balance between privacy preservation and data utility remains a critical challenge.

To address this issue, we introduce a novel bilevel optimization framework for the publication of private datasets, where the upper-level task focuses on data utility and the lower-level task focuses on data privacy. In the upper-level task, a discriminator guides the generation process to ensure that perturbed latent variables are mapped to high-quality samples, maintaining fidelity for downstream tasks. In the lower-level task, our framework employs local extrinsic curvature on the data manifold as a quantitative measure of individual vulnerability to MIA, providing a geometric foundation for targeted privacy protection. By perturbing samples toward low-curvature regions, our method effectively suppresses distinctive feature combinations that are vulnerable to MIA. Through alternating optimization of both objectives, we achieve a synergistic balance between privacy and utility. Extensive experimental evaluations demonstrate that our method not only enhances resistance to MIA in downstream tasks but also surpasses existing methods in terms of sample quality and diversity.

\end{abstract}




\maketitle

\section{Introduction}
The training of machine learning models is fundamentally dependent on the availability of high-quality datasets. However, direct sharing of sensitive private data - particularly in critical domains such as healthcare, financial services, and personalized recommendation systems - introduces substantial privacy risks, as it may lead to unauthorized exposure of personal information, potentially resulting in privacy violations or misuse of sensitive data. To mitigate these risks, many private datasets have been released after undergoing various preprocessing and de-identification techniques. For instance, the Netflix Prize dataset\cite{bennett2007netflix} obfuscates the timestamps of user ratings and removes all identifiers to safeguards the privacy of their customer base. Similarly, the ADNI dataset\cite{neeraj_kumar_2021} de-identifies participants' demographic and genetic information to ensure data anonymity and compliance.

Although de-identification techniques prevent direct privacy leakage by removing explicit identifiers, they fail to address the implicit privacy risks arising from the model's memorization of data features. Even when identifiable attributes are removed, attackers can still exploit learned patterns and distinctive feature combinations, which may serve as unique identifiers. As a result, simple attribute obfuscation is insufficient to defend against advanced privacy threats, such as membership inference attacks (MIA)\cite{shokri2017membership}, where attackers infer whether a specific sample was part of the training data based solely on the model’s outputs. Furthermore, for data publishers, the challenge extends beyond defending against attacks, as the uncertainty of downstream tasks complicates data release. For example, in a case where a dataset owner publishes image data, the downstream tasks could involve generation, classification, or other analyses, and this unknown purpose may render region masking technique ineffective, especially for generation tasks. Therefore, it is crucial to develop responsible data release methods that protect privacy while ensuring utility for downstream use.

Current research efforts in privacy-preserving data publishing have primarily focused on three key strategies: data perturbation, generalization, and synthetic data generation. For example, differential privacy (DP)\cite{abadi2016deep} offers strong theoretical privacy guarantees by adding noise to the data, but this noise can reduce data quality, limiting its practicality for complex tasks. Similarly, k-anonymity\cite{sweeney2002k} groups data to prevent individual identification, but this can result in a loss of information diversity. So, to address this limitation, l-diversity\cite{machanavajjhala2007diversity} requires that sensitive attribute values within each equivalence class be diverse. However, while this method can prevent homogeneity attacks, implementing it with high-dimensional datasets, such as image datasets, remains challenging. PATE-GAN\cite{jordon2018pate} relies on a system voting involving multiple teacher discriminators which guide a student discriminator. The method creates a privacy-preserving GAN, but the generated samples seldom comprehensively cover the original data distribution, leading to limited diversity. Plus, the injected noise usually degrades the fidelity of the image details. PrivBayes\cite{zhang2017privbayes} models attribute dependencies using Bayesian networks and adds noise at each learning step to achieve DP, but this method is restricted to tabular data. Other methods that integrate DP into synthetic models to generate privacy-preserving data, such as WS-GAN\cite{chen2020gs}, PrivBayes\cite{zhang2017privbayes} and DPDM\cite{dockhorn2022differentially}, also suffer from reduced data quality due to noise injection. Generally, although these methods offer certain trade-offs between privacy and utility, they often introduce excessive noise or apply overly generalized transformations, which reduce data resolution, diversity, and adaptability for various downstream tasks, and ultimately undermine data utility. Thus, the core challenge is to strike an improved balance between utility and privacy.

Our method addresses this issue through a bilevel optimization framework. The upper-level task focuses on maximizing the utility of the generated data, while the lower-level task targets privacy preservation by perturbing vulnerable data points. To achieve the upper-level objective, a discriminator is integrated to enhance latent space exploration and ensure high-quality generated samples. To achieve the lower-level objective, we identify that certain samples in a dataset are inherently more vulnerable to MIA because they are more likely to be memorized by the model rather than generalized, such as those with unusual feature combinations, outliers, or points near decision boundaries. Leveraging the extrinsic curvature information embedded in the data manifolds, we pinpoint these vulnerable data points and apply targeted perturbations to mitigate the risk of MIA. Specifically, we employ a Riemannian Variational Autoencoder (RVAE) as the backbone model, which simultaneously functions as a generative model and a manifold learning tool. The RVAE captures the intrinsic structure of the data manifolds and provides a Riemannian metric for curvature computation, enabling efficient identification of vulnerable regions. Guided by this curvature information, geodesic interpolation is used to perturb samples toward low-curvature regions, effectively reducing the risk of MIA. Since discriminator-guided updates to the decoder influence the Riemannian metric and create interdependencies among network components, we use alternating optimization to manage the coupling process, ensuring a balanced and optimized model that simultaneously achieves high generation quality and precise perturbation.

The main contributions of this work can be summarized as follows:
\begin{itemize}
\item We propose a novel bilevel framework that significantly improves the trade-off between privacy and utility in data publication. To the best of our knowledge, this is the first approach that explicitly balances generation quality and MIA defenses through bilevel optimization.

\item We propose a curvature-guided geodesic perturbation method that leverages the intrinsic features of the data manifold to guide perturbations, selectively avoiding vulnerable regions and effectively reducing the success of MIA, thus protecting individual privacy.

\item We generate high-quality synthetic samples that preserve data diversity while suppressing distinctive feature combinations, outperforming traditional techniques. 

\end{itemize}

\section{Literature Review}
This section provides a review of relevant work across two spheres of study: MIA and MIA defenses at the data publishing stage.

\subsection{Membership Inference Attacks} 
The primary goal of MIA is to deduce whether or not a given data record was used to train a target model. Based on their underlying principles, MIAs can be divided into four primary categories: binary-classifier based attacks\cite{shokri2017membership}\cite{salem2018ml}, evaluation metrics-based attacks, differential comparisons-based attacks\cite{hui2021practical}, and generalization gap and model metrics-based attacks\cite{bentley2020quantifying}. Attacks based on evaluation metrics can be further divided into several subcategories, including attacks based on prediction confidence \cite{song2019privacy}, those based on prediction loss \cite{carlini2022membership}, those based on prediction labels \cite{choquette2021label}, and those based on prediction entropy \cite{song2021systematic}. 

The success of an MIA is influenced by various factors, but generally, success is primarily due to issues with the model itself or vulnerabilities in the dataset. In terms of the model, overfitting is a key concern, as overfit models are more likely to memorize and leak features of the training samples\cite{song2020membership}. A larger generalization gap also correlates to an increased vulnerability to attacks\cite{li2021membership}. Additionally, privacy vulnerabilities in training algorithms, such as potential information leakage in stochastic gradient descent under certain conditions, also pose risks\cite{nasr2019comprehensive}. Turning to the datasets, long-tail distributions may make rare samples more identifiable\cite{truex2019effects}. High-risk data points and unique features also increase the likelihood of successful inference\cite{song2021systematic}\cite{duddu2021shapr}.

\subsection{Defences against MIA} 

Methods for defending against MIA can be divided into data publishing and model training methods, depending on the stage at which they are implemented. Our work focuses defenses implemented during the data publishing stage. These privacy-preserving solutions can be classified into three principal schemes: data perturbation, generalization, and synthetic data generation.

\textbf{Data perturbation} is a commonly used privacy protection technique, which works by adding random noise to data to obscure sensitive information. The idea is to make it more difficult for attackers to reconstruct the original data. DP\cite{abadi2016deep} is a typical example of this technique. It ensures the privacy of individual information by adding noise to data or model outputs while retaining the overall statistical properties of the dataset. However, the introduction of noise can significantly reduce the utility of the data, especially when dealing with tasks requiring high precision in analysis.

\textbf{Generalization} in data publishing enhances privacy by reducing the identifiability of individuals through methods like aggregation, grouping, or obfuscation. Pixelation and blurring techniques lower the resolution of images or data, making sensitive information harder to identify\cite{gross2009face}\cite{vishwamitra2017blur}. K-anonymity\cite{sweeney2002k} and its extensions, l-diversity\cite{machanavajjhala2007diversity} and t-closeness\cite{li2006t}, generalize individual data into groups with similar features, to prevent unique identification, while Manifold Mixup\cite{verma2019manifold} is a generalization technique that maps data points onto a manifold and linearly interpolates different samples and labels in the feature space, generating more generalized and harder-to-infer samples. This encourages the model to learn more robust decision boundaries and prevents overfitting. However, excessively reducing individual details in the data can lead to a significant loss of accuracy and information. For instance, pixelation and blurring obscure image details, K-anonymity reduces dataset diversity, and Manifold Mixup's linear interpolation in high-dimensional feature space may generate synthetic samples that deviate from the data manifold. As such, subsequent processing may fail to capture the intrinsic geometric structures and nonlinear features of the data.

\textbf{Synthesizing new data} similar to the original data distribution can be done through synthesis techniques such as GANs or VAEs. These methods protect the privacy of individuals by not directly publishing any real data \cite{yoon2020anonymization}\cite{gross2023differentially}. However, generative models rely on accurately capturing the data distribution. GANs can suffer from mode collapse, producing similar samples and reducing diversity. Constrained by the simplicity of the prior, data generated by a VAE may appear slightly blurry or lack detail. These could potentially affect the performance of downstream models in practical applications. Hence, several synthetic privacy-preserving frameworks have been proposed. For example, PATE-GAN\cite{jordon2018pate} combines the PATE framework with GANs to generate differentially-private synthetic data, making this framework suitable for releasing sensitive data. However, this method requires training multiple teacher models, which leads to high computational costs and a rather complex training process. PrivBayes\cite{zhang2017privbayes} models data correlations using Bayesian networks and releases synthetic data through a DP mechanism, which effectively handles high-dimensional data but is limited to tabular data. Its ability to process images and other types of data is severely limited. SoK\cite{hu2024sok} provides a systematic analysis of the application of statistical and deep learning methods in privacy-preserving data synthesis, offering guidance for different scenarios, but the added noise does impact data quality. Dockhorn et al. proposed DPDM\cite{dockhorn2022differentially}, which integrate DP into diffusion models. Although the diffusion model as the backbone has strong generative capabilities, the noise introduced by the DP mechanism still degrades the quality of the generated data.

\section{Preliminaries}

Our model is built upon a bilevel optimization framework, utilizing a RVAE as its backbone to enable geodesic perturbations. This section
introduces the foundational concepts and research findings that
underpin our work.

\subsection{Bilevel Optimization}

Bilevel optimization involves two nested optimization tasks: a upper-level problem and a lower-level problem. The lower-level problem serves as a constraint for the upper-level problem, creating a hierarchical dependency between them. It has been applied in various domains, such as meta-learning \cite{finn2017model} and energy system optimization \cite{wogrin2020applications}. Formally, given upper-level variables \( x \in \mathbb{R}^m \) and lower-level variables \( y \in \mathbb{R}^n \), the tasks are defined as:

\begin{equation}
\min_{x \in \mathbb{R}^{d_x}} F(x) = f(x, y^*(x)), \quad
\text{s.t. } y^*(x) = \arg\min_{y \in \mathbb{R}^{d_y}} g(x, y).
\label{eq:bilevel_tasks}
\end{equation}

Here, \( g(x, y) \) and \( f(x, y^*(x)) \) are the lower- and upper-level objectives, respectively. In our framework, the upper-level task maximizes data utility, while the lower-level task enhances privacy through targeted perturbations.

\subsection{Riemann VAE} 

Traditional VAEs\cite{kingma2013auto} use a Gaussian distribution as a prior, which limits their expressiveness because the latent structure of the data is unlikely to fit a predefined Gaussian prior. The Riemannian VAE(RVAE)\cite{arvanitidis2017latent}\cite{kalatzis2020variational}\cite{chadebec2022geometric} introduces the concept of Riemannian geometry to capture the intrinsic complexities and local variations in the data. By leveraging this geometry, the latent space is modeled as a curved manifold that adapts to the underlying structure of the data, allowing for a better representation and higher sample quality.

In the RVAE framework, the generative distribution is parameterized as follows:
\begin{align}
    f(z) &= \mu(z) + \sigma(z) \odot \epsilon, \quad \epsilon \sim \mathcal{N}(0, I_M) \label{eq:vae_sampling}
\end{align}
where \( z \in \mathbb{R}^d \) denotes the latent variable, \( \mu: Z \to \mathbb{R}^M \) and \( \sigma: Z \to \mathbb{R}_+^M \) represent the mean and standard deviation, respectively. Both \( \mu \) and \( \sigma \) are parameterized by separate neural network structures. Notably, \( \sigma \) employs radial basis functions (RBFs) to provide a more accurate and stable local manifold structure and a better estimation of the uncertainty present.

RVAE introduces a pullback metric as a Riemannian metric on the latent space \( Z \). The pullback metric at a point on an embedded manifold is defined as the standard Euclidean metric transferred to the parameter space via a local coordinate mapping, defined as:
\begin{align}
    G(z) &= J_{\mu}(z)^T J_{\mu}(z) + J_{\sigma}(z)^T J_{\sigma}(z) \label{eq:pullback_metric}
\end{align}
where \( J_{\mu}(z) \) and \( J_{\sigma}(z) \) are the Jacobians of the mappings \( \mu(z) \) and \( \sigma(z) \), respectively. This metric captures the intrinsic geometry of the manifold by relating the local geometry of the data space to that of the latent space. Consequently, the Riemannian metric provides a way to measure curvatures on the latent manifold.

\subsection{Manifold Curvature}

Previous curvature work has focused more on the curvature of decision boundaries or the neural network parameter manifold\cite{fawzi2018empirical}\cite{yu2018curvature}. Ravikumar et al. discovered that input loss curvature can be used for membership inference, providing new insights into the gap between the training and test sets\cite{ravikumar2024curvature}. However, we are concerned with the extrinsic curvature of the manifold itself. High curvature regions represent the "sharp" parts of the data distribution, reflecting the local nonlinear features of the data, corresponding to more unique or rare samples, or samples with greater noise. These regions may also lie near decision boundaries\cite{kaul2019riemannian}\cite{kaufman2023data}, increasing the likelihood of misclassification and making it harder for models to learn robust representations. Therefore, models may have difficulty generalizing to new samples in areas of high curvature. Few models currently consider extrinsic curvature, with CurvVAE\cite{rhodes2022learning} being one of the few that regularizes the extrinsic curvature of the learned manifold in the data space to avoid model overfitting. 

Because the computation of the Hessian matrix and the second fundamental form is too computationally intensive, and the relative curvature of different regions is sufficient for our work, we applied an approximation of the extrinsic curvature. For more details, please refer to the Subsubsection.~\ref{sec:metho_curvature}.

\subsection{Geodesic Interpolation} 
\label{sec:geodesic_interpolation}

A geodesic is the shortest path between two points on a Riemannian manifold. Geodesic interpolation is a method of creating intermediate points between two points on a manifold, where these intermediate points lie on the geodesic connecting the two points. Because it preserves the geometric structure of the manifold, interpolation along the geodesic produces more natural and coherent results, superior to linear interpolation\cite{syrota2024decoder}\cite{struski2023feature}.

Our framework follows a cubic spline interpolation method to compute the geodesics. More specifically, spline curves are generated for each pair of starting and ending points, and the changes in mean and variance are calculated to assess the smoothness of the curve. The curve energy can be expressed as:
\begin{align}
    \text{Energy}_{\text{curve}} &= \frac{1}{2} \sum \left( \|\Delta \mu\|^2 + \|\Delta \sigma\|^2 \right) \label{eq:energy_curve}
\end{align}
Here, $\Delta \mu$ and $\Delta \sigma$ represent the changes in the mean and standard deviation, respectively. Smooth geodesics are derived by optimizing the curve parameters to minimize the curve energy.

\section{Methodology}

The framework of our model is illustrated in Fig. \ref{fig:framework}, including the pipeline and our model architecture. The section begins with a summary of the design and components, followed by the optimization details of the upper-level and lower-level tasks. Then, the bilevel optimization process and algorithm flow are introduced.

\subsection{Overview}

\begin{figure*}[!htbp]
    \centering
    \includegraphics[width=1\linewidth]{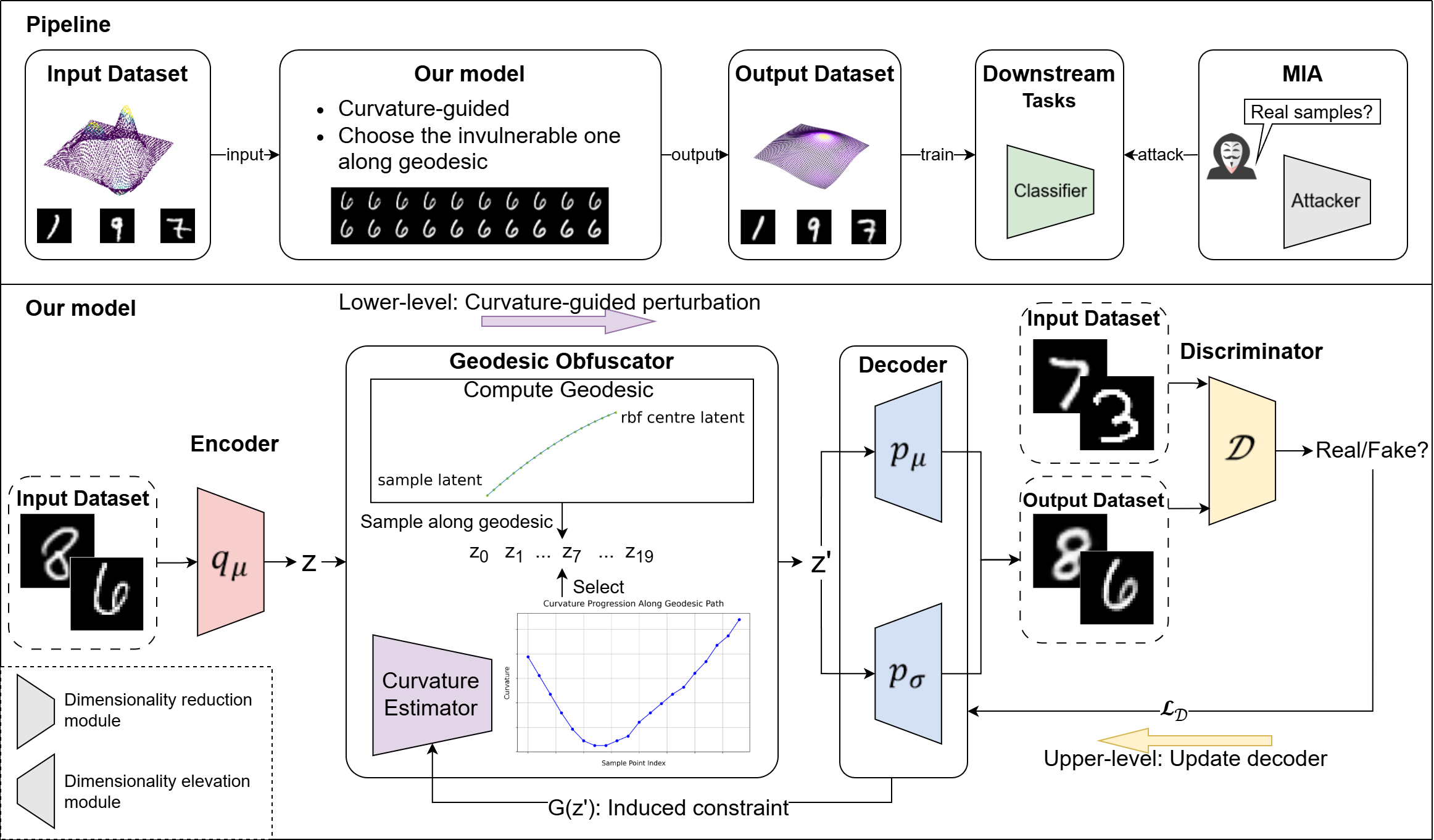}
    \caption{Overview of the Framework: Experimental Pipeline and Our Proposed Model}
    \label{fig:framework}
\end{figure*}

The general pipeline and model design are shown in Fig. \ref{fig:framework}. Our goal is to generate a new dataset from the original private dataset, which can be used for downstream tasks without allowing attackers to infer the original private data from the downstream model. To balance data privacy protection and data utility preservation, we construct a bilevel optimization framework:

\textbf{Upper-Level Task}: The goal of the upper-level task is to improve the reconstruction quality of geodesically perturbed points, ensuring they closely resemble the samples from the original dataset.

\textbf{Lower-Level Task}: The lower-level task aims to improve the estimation accuracy of the curvature estimator, thereby ensuring that perturbations move towards regions of low curvature, which guides the selection of geodesic sampling points.

Our framework consists of an RVAE, a geodesic obfuscator, and a discriminator. The RVAE, which includes an encoder and a decoder, serves as the backbone. It not only performs image reconstruction but also facilitates manifold learning, providing a pullback metric for subsequent curvature calculations. During the upper-level optimization, the discriminator facilitates latent space exploration while preserving the quality of reconstructed images under perturbation. For simplicity, we refer to the combination of the RVAE and discriminator as RVAE-GAN. During the lower-level optimization process, the geodesic obfuscator applies curvature-guided perturbations to the latent variables. It comprises two key components: an optimizable curvature estimator and a unit for perturbations along geodesics. The extrinsic curvature of the data manifold serves as an indicator of vulnerability, and a detailed mathematical analysis of the relationship between high-curvature regions and vulnerability to MIA is provided in Appendix \ref{proof}. Based on this insight, the curvature estimator functions similarly to a regression model, guiding the perturbations. We assume that curvature does not vary significantly within the relevant domain, allowing a well-trained estimator to eliminate the need for repeatedly computing the local curvature of sampled points during geodesic perturbations. A fundamental aspect of our framework is the interdependence between the upper and lower-level optimizations. The decoder updates in the upper-level task modify the pullback metric, which then influences the lower-level optimization through its role as an induced constraint. This constraint effectively couples both optimization processes, creating a cohesive computational architecture. To evaluate the effectiveness of the synthetic dataset, we assess its performance on downstream tasks using a classifier and conduct a black-box loss-based MIA.

The following subsections detail our framework’s functional modules and training process, structured around the upper-level and lower-level tasks, followed by an algorithm summary. Subsection \ref{sec:RVAE-GAN} covers the upper-level task, introducing RVAE-GAN as the model backbone and its training process to enhance generation quality. Subsection \ref{sec:Geodesic_Obfuscator} focuses on the lower-level task, describing the geodesic obfuscator, which includes the curvature estimator and geodesic interpolation unit for data perturbation. Subsection \ref{sec:BLO} demonstrates how the optimization of the two levels influences each other and presents the alternating training strategy. Finally, subsection \ref{sec:Algorithm} presents the complete training process and algorithm flow.

\subsection{Upper-Level Task: RVAE-GAN for High-quality Reconstruction}
\label{sec:RVAE-GAN}

To preserve the utility of reconstructed images, we adopt RVAE as the backbone and incorporate a discriminator to further enhance fidelity. With its flexible prior, RVAE serves as a powerful generative model. However, insufficient latent space exploration can compromise the quality of geodesic interpolation. To mitigate this, we introduce a discriminator that helps maintain the generated samples within the data distribution while encouraging a more thorough exploration of the latent space.

During the pre-training phase, updates for the RVAE and the discriminator are conducted in an alternating manner. The RVAE updates are split into two stages: the $\mu$ stage and the $\sigma$ stage.

In the $\mu$ stage, the focus is on optimizing the mean parameters of both the encoder and the decoder. Here, the RVAE is optimized using the mean loss function $\mathcal{L}_{\mu}$, while the discriminator is simultaneously trained using both generated and real images. The discriminator loss from the generated images contributes to optimizing the mean parameters of the encoder and decoder, ensuring that these components learn robust representations.

In the $\sigma$ stage, the primary goal is to optimize the standard deviation parameters of the decoder as well as the prior mean in the latent space. During this stage, the RVAE is updated based on the variance loss function $\mathcal{L}_{\sigma}$, while the discriminator continues to be trained on real and generated images. Here, the discriminator loss from generated images is used to update the decoder's standard deviation and the prior mean, refining the latent space distribution to enhance the model’s generation quality.

The loss functions for the $\mu$ and $\sigma$ stages are as follows:

\begin{align}
    \text{$\mu$ Loss: } & \mathcal{L}_{\mu} = -\mathbb{E}_{q(z|x)}[\log p(x|z)] \\
    \text{$\sigma$ Loss: } & \mathcal{L}_{\sigma} = \mathbb{E}_{q(z|x)}\left[-\log p(x|z) \right. \nonumber \\
    & \quad \left. + \beta \cdot D_{KL}(q(z|x) \| p(z))\right] 
\end{align}
\noindent where \( x \) represents the observed data sample, and \( z \) denotes the latent variable. The term $D_{KL}(q(z|x) \| p(z))$ is the Kullback-Leibler (KL) divergence between the variational posterior $q(z|x)$ and the prior $p(z)$.

The KL divergence and the log pdf of a Brownian motion (BM) transition kernel are given by:

\begin{align}
    D_{KL}(q(z|x) \| p(z)) = \mathbb{E}_{q(z|x)}\left[\log_{\text{bm}} q(z|x) - \log_{\text{bm}} p(z)\right].
\end{align}

\begin{equation}
    \log_{\text{bm}} q(z|x) = -\frac{d}{2} \log(2 \pi \sigma_z^2) + \frac{1}{2} \log \left| \det G(z) \right|
    - \frac{1}{2} \log \left| \det G(\mu_z) \right| - \frac{l^2(z, \mu_z)}{2 \sigma_z^2}.
\end{equation}

\begin{equation}
    \log_{\text{bm}} p(z) = -\frac{d}{2} \log(2 \pi) + \frac{1}{2} \log \left| \det G(z) \right|
    - \frac{1}{2} \log \left| \det G(\mu_{\text{prior}}) \right| - \frac{l^2(z, \mu_{\text{prior}})}{2}.
\end{equation}

where \( d \) is the dimensionality of the manifold; \( l^2 \) is the squared distance between points in the manifold; and \(G\) is the pullback metric as defined in \ref{eq:pullback_metric}. Here, \( \mu_{\text{prior}} \) represents the prior mean, and \( \mu_z \) denotes the posterior mean, both of which are learned through the training process.

The discriminator step loss is given by: 
\begin{equation}
\mathcal{L}_D = \mathbb{E}[D(f(z))] - \mathbb{E}[D(x)] + \lambda \cdot GP,
\label{eq:loss_d}
\end{equation}
where \( D \) represents the output of the discriminator, with inputs being either generated samples \( f(z) \) or real samples \( x \); \( f(z) \) denotes the output of the generator (RVAE), where \( z \) is a latent variable; and \( GP \) is the gradient penalty term for regularization, weighted by \( \lambda \). 

The generator (RVAE) step loss is given by: 
\begin{equation}
\mathcal{L}_G = -\mathbb{E}[D(f(z))],
\end{equation}
where \( f(z) \) represents the output of the generator (RVAE) and \( D(f(z)) \) indicates the discriminator's judgment on the generated samples.

In addition to functioning as a generative model, the RVAE effectively captures the intrinsic properties of the data manifold, enabling subsequent curvature computation. Furthermore, its RBF kernels facilitate privacy-preserving interpolation by serving as endpoints for geodesics. We will discuss it in the following subsection.

\subsection{Lower-Level Task: Geodesic Obfuscator for Curvature-Guided Perturbation}
\label{sec:Geodesic_Obfuscator}

Our geodesic obfuscator achieves the goal of perturbing data to protect individual privacy. It primarily consists of two components: a trainable curvature estimator and a geodesic interpolation unit. The curvature estimator takes latent variables as input and outputs local extrinsic curvature estimates, functioning similarly to a regression network. The second component, the geodesic interpolation unit, selects the RBF center with the largest contribution for each latent variable sample, calculates the geodesic, then collects samples along this geodesic, and ultimately returns a perturbed latent variable with an appropriate curvature. Next, we will describe these two modules in detail.

\subsubsection{Curvature estimator.}
\label{sec:metho_curvature}

In the framework, a computationally efficient finite difference method approximates the extrinsic curvature. Although this may not be as precise as calculations based on the full second fundamental form, it does provide sufficient geometric information for comparing the relative curvatures across different regions. The Riemannian metric $G(z)$ and its eigenvalues are computed for each point $z$. Then small perturbations are applied to $z$ and the rate of change of these eigenvalues is calculated. The norm of these change rates serves as our curvature estimate, capturing the degree of metric tensor variation in the neighborhood of $z$. This method allows for efficient curvature estimation in high-dimensional latent spaces without the need for computationally expensive Hessian calculations.

Mathematically, the curvature estimate $K(z)$ is defined as:

\begin{align}
    K(z) = \|\nabla\lambda(G(z))\|
\end{align}

where $\lambda(G(z))$ represents the eigenvalues of $G(z)$, and $\nabla\lambda(G(z))$ is approximated using finite differences:

\begin{align}
    \nabla\lambda(G(z)) \approx \frac{\lambda(G(z+\epsilon)) - \lambda(G(z))}{\epsilon}
    \label{eq:curvature_estimate}
\end{align}

Here, $\epsilon$ represents a small perturbation in the latent space.

Training the curvature estimator is done through a mean squared error (MSE) loss function:

\begin{align}
    \mathcal{L}_{\text{curv}} = \mathbb{E}_{z}\left[\left(\hat{K}(z) - K(z)\right)^2\right]
\end{align}

where $\hat{K}(z)$ is the curvature predicted by our estimator model, and $K(z)$ is the approximated curvature value obtained using our finite difference method.

\subsubsection{Perturbation strategy.}

We employ a fine-grained geodesic perturbation method, offering more precise control than linear interpolation. The current point \( z \) is treated as the starting point of the geodesic, and the endpoint of the geodesic is selected as the point that contributes the most among the RBF centers of the RVAE. The geodesic is computed by optimizing a cubic spline, as explained in subsection \ref{sec:geodesic_interpolation}, and \( n \) equidistant sample points \( \{ z_1, z_2, \dots, z_n \} \) are taken along the path. Curvature is estimated at each of these \( n \) sample points as \( \hat{K}(z_i) \), and the point with the lowest curvature before the high curvature regions are reached is selected as the endpoint for the geodesic perturbation. In rare cases, if the most contributive RBF center lies in the region of another category, geodesic interpolation might cross the category boundary, leading to invalid labels. By restricting interpolation to avoid crossing the areas of highest curvature along the geodesic, we limit boundary-crossing, which helps to maintain label integrity. The result point \( z' \) is determined by:

\begin{align}
    z' = \arg \min_{z_i \in \{z_0, z_1, \dots, z_{i_{\text{max}}}\}} \hat{K}(z_i), \label{eq:min_z}
\end{align}

where \( \hat{K}(z_i) \) is the estimated curvature and \( i_{\text{max}} \) is the index of the point with the highest curvature along the geodesic.

\subsection{Bilevel Optimization}
\label{sec:BLO}

\begin{figure*}[!htbp]
    \centering
    \includegraphics[width=0.9\linewidth]{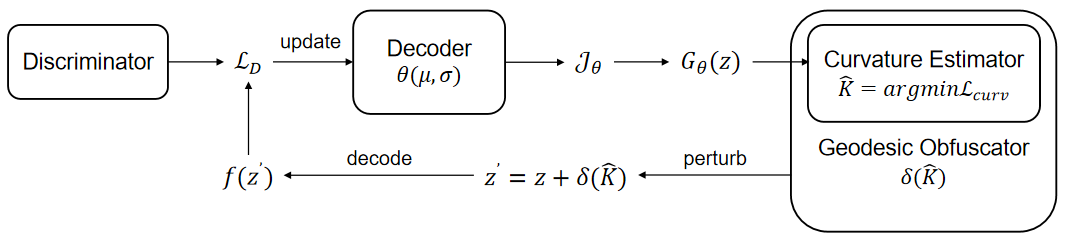}
    \caption{Coupling Process in Bilevel Optimization}
    \label{fig:bilevel}
\end{figure*}

During the optimization of the upper-level task, the discriminator loss updates the decoder, leading to changes in the Jacobian matrix. These changes directly influence the pullback metric, which acts as an induced constraint that governs the curvature estimations in the lower-level task, thereby fundamentally altering the perturbation results. This interdependent process, illustrated in Fig. \ref{fig:bilevel}, forms a coupled loop that we frame as a bilevel optimization problem. Solving this problem allows the model to achieve an optimal balance between generation quality and effective perturbation. The optimization objectives can be expressed in the following form:

\begin{align}
    &\theta = \arg\min \, \mathcal{L}_D(f(z + \delta(\hat{K}))) \\
    &\text{s.t.} \quad \hat{K} = \arg\min \, \mathcal{L}_{\text{curv}}(\hat{K}, \theta)
\end{align}

where the parameter $\theta$ is the decoder's parameters, including the $\mu$ and $\sigma$ networks. The function $f$ is the generative function, as referenced in Equation \ref{eq:vae_sampling}. $\delta(\hat{K})$  represents the geodesic perturbation process. The \(\mathcal{L}_{\text{curv}}\) is determined by \(\|\nabla\lambda(G(z))\|\), which is influenced by the Jacobian matrix of the decoder network. This is denoted as a function of $\theta$, as referenced in Equation \ref{eq:pullback_metric} and \ref{eq:curvature_estimate}. 

In the bilevel optimization phase, an alternating training method is used, which optimizes each module separately across four steps. These steps sequentially update different components of the model as follows:

\begin{enumerate}
    \item Similar to the pre-training phase, the RVAE is trained without geodesic perturbations to prevent excessive updates that could deviate from the original manifold.
    \item The discriminator is updated using original data for real images and geodesic-perturbed reconstructions for fake images, continuously enhancing its guiding ability.
    \item The decoder is updated using the discriminator loss to improve its exploration of the latent space.
    \item The curvature estimator is updated to maintain its accuracy under the decoder's subtle changes.
\end{enumerate}

Through this coordinated interaction process, we refine geodesic perturbations and enhance the model’s generative capabilities.

\subsection{Algorithm}
\label{sec:Algorithm}

\begin{algorithm}[htbp]
    \caption{Bilevel Optimization with Curvature-guided Perturbation}
    
    \SetKwInOut{Input}{Input}
    \SetKwInOut{Output}{Output}
    
    \textbf{Phase 1: Pre-train RVAE-GAN ($\mu$ step and $\sigma$ step)} \\
    
    \textbf{Step 1: $\mu$ step} \\
    Optimize RVAE to minimize $\mathcal{L}_{\mu}$ \\
    Optimize discriminator to minimize $\mathcal{L}_{\text{D}}$ \\
    Optimize RVAE to minimize $\mathcal{L}_{\text{G}}$ \\
    
    \textbf{Step 2: $\sigma$ step} \\
    Optimize RVAE to minimize $\mathcal{L}_{\sigma}$ \\
    Optimize discriminator to minimize $\mathcal{L}_{\text{D}}$ \\
    Optimize RVAE to minimize $\mathcal{L}_{\text{G}}$ \\

    \textbf{Phase 2: Pre-train Curvature Estimator} \\
    Optimize Curvature Estimator to minimize $\mathcal{L}_{\text{curv}}$ \\

    \textbf{Phase 3: Bilevel Optimization} \\
    \ForEach{iteration}{
        \textbf{Step 1: RVAE step} \\
        $z \gets g(x)$ \\
        $x' \gets f(z)$ \\
        Minimize $\mathcal{L}_{\text{ELBO}}$ for RVAE \\

        \textbf{Step 2: Discriminator step} \\
        $z \gets g(x)$ \\
        $z' \gets z + \delta(K)$ \\
        $x' \gets f(z')$ \\
        Minimize $\mathcal{L}_{\text{D}}$ using both fake images $x'$ and real images \\

        \textbf{Step 3: Decoder step} \\
        Optimize decoder to minimize $\mathcal{L}_{\text{G}}$ \\

        \textbf{Step 4: Curvature estimator step} \\
        Optimize curvature estimator to minimize $\mathcal{L}_{\text{curv}}$ \\
    }
    \label{algorithm}
\end{algorithm}

The complete algorithm flow is presented in Algorithm \ref{algorithm}. The training process is divided into three phases: pre-training the RVAE-GAN, pre-training the curvature estimator, and the bilevel optimization process with added geodesic perturbation.

\section{Experiments}

This section provides a detailed description of the experiments used to evaluate the effectiveness and functionality of the proposed framework. We first outline the experimental setup, including the datasets, baseline models, and training hyperparameters. Then, we present the evaluation metrics employed to assess both privacy and utility. Next, we compare the performance of different models on these metrics. Subsequently, we provide visualizations of the latent space and compare different interpolation strategies. Finally, we analyze the extrinsic curvature and its relationship with sample vulnerability.

\subsection{Setup} 
\subsubsection{Datasets.}

Experiments were conducted on the MNIST \cite{lecun1998gradient}, Fashion-MNIST (FMNIST) \cite{xiao2017fashion}, Kuzushiji-MNIST (KMNIST) \cite{clanuwat2018deep}, small notMNIST\cite{bulatov_notmnist}, Math Symbols\cite{amola_math_symbols}, Kannada-MNIST(KANMNIST)\cite{prabhu_kannada_mnist}, Afro-MNIST\cite{wu_afro_mnist} and dSprites\cite{matthey_dsprites} datasets. MNIST consists of grayscale images of handwritten digits (0-9), while FMNIST includes grayscale images of clothing items such as T-shirts, pants, and shoes. KMNIST features classical Japanese characters across ten classes, and small notMNIST consists of hand-sculpted grayscale images of letters from A to J derived from various fonts. The Math Symbols dataset contains images of digits (0-9) and four mathematical operators (addition, subtraction, multiplication, division), from which we selected a subset for our experiments containing only the digits 0-5 and the mathematical operators. KANMNIST is a new handwritten digits dataset for the Kannada script. Afro-MNIST is a collection of synthetic MNIST-style datasets representing four orthographies from Afro-Asiatic and Niger-Congo languages, among which we only used the Ethiopia dataset for our experiments. The dSprites dataset is a synthetic dataset of 2D shapes with four controllable factors of variation (shape, scale, rotation, and position) that was created for disentanglement research. To add data diversity, we assigned colors based on the scale attribute and used shape as the target label for a downstream classification task. Additionally, we included OCTMNIST \cite{medmnistv2}, which comprises retinal optical coherence tomography (OCT) images reformatted into the MNIST-style format, providing a medical imaging perspective to our evaluation framework.

To more thoroughly evaluate MIA defenses, we processed each dataset into varying degrees of class imbalance to amplify their vulnerabilities and assess the effectiveness of each model's defense performance. The datasets were processed as follows. The training sets of MNIST, KMNIST, FMNIST, and KANMNIST were down-sampled to create class-imbalanced distributions (specifically, classes 6, 7, 8, and 9 for MNIST, and the last three classes for FMNIST, KMNIST, and KANMNIST). We down-sampled to 10\% of the head class size, leaving the test sets unchanged. This simulated a scenario where the attacker was more focused on minority class samples. For the Math Symbols, Afro-MNIST, and notMNIST datasets, both the training and test sets were down-sampled to form class-imbalanced distributions using the same down-sampling approach as described above. Here, the multiply and subtract classes for Math Symbols and the last three classes for Afro-MNIST and notMNIST were the minority classes. Lastly, the dSprites dataset retained only 20\% of the data without introducing any class imbalance, simulating a robust original dataset. The OCTMNIST dataset was utilized in its original form, with both training and test sets remaining unaltered.

\subsubsection{Baseline methods.}

The following baseline methods we used for comparison:

\begin{itemize}
    \item \textbf{Traditional image blurring techniques (pixelation and blur):} These methods obscure image details by reducing the resolution (pixelation) or applying a smoothing filter (blur) to degrade visual clarity.
    
    \item \textbf{K-anonymity method:} This method uses clustering, where all samples in a cluster are replaced by the cluster center, ensuring that individual data points are indistinguishable within the group.
    
    \item \textbf{VAEGAN-DP:} Following \cite{xie2018differentially} and \cite{chen2020gs}, we reproduced the DP-GAN with a VAE generator for conditional image-to-image generation, applying DP only to the generator. This creates a model similar to our backbone but with DP-based perturbations.

    \item \textbf{DPDM:} Following \cite{dockhorn2022differentially}, we implemented the DPDM that applies DP-SGD during training. It leverages noise multiplicity to reduce gradient variance, improving generation quality while maintaining privacy guarantees.
\end{itemize}

\subsubsection{Hyperparameters.}

During the pre-training phase of RVAE, the first 100 epochs were dedicated to the $\mu$-stage, followed by 100 epochs for the $\sigma$-stage, with both optimizers having a learning rate of $1\times10^{-3}$. Since the discriminator learns faster, it was updated every 50 iterations, and the discriminator loss was used to guide the updates of RVAE. The learning rate of the discriminator was lower, set at $1\times10^{-6}$. Subsequently, the curvature estimator was optimized for 50 epochs with a learning rate of $1\times10^{-4}$. Finally, the bilevel optimization phase lasted for 5 epochs, with fine-tuning for both the RVAE and the discriminator at a learning rate of $1\times10^{-5}$. The number of sampling points along the geodesic was set to 20.

Our comparative experiments relied on 3x3 pixelation and a Gaussian blur filter with a radius of 1.5. For K-anonymity, we set K=10 and performed K-means clustering on the data, with the number of cluster centers set to 1000. When the number of samples within a cluster exceeded K, we replace all data points in the cluster with the nearest data point to the cluster center and its corresponding label. We implemented a VAE-GAN model with VAE as the generator, applying DP through the PrivacyEngine\cite{opacus_privacy_engine} during the training process of the VAE, adding noise while also implementing gradient clipping. The noise multiplier was set to 1.1, and the max grad norm for gradient clipping was set to 1.0. We also implemented DPDM based on PrivacyEngine, setting the noise multiplier to 0.7, the clipping bound to 2.0, and the noise multiplicity parameter to 3.

\subsection{Metrics} 

We evaluated the model's performance from the standpoints of both privacy and utility.

\subsubsection{Privacy metrics.}

To evaluate the resistance of the generated dataset to MIA, we adopt a loss-based MIA method, using the MembershipInferenceBlackBox method from the adversarial-robustness-toolbox \cite{nicolae2018art} to attack the downstream classifier. The downstream classifier is trained on the reconstructed dataset, while the attack model uses the original training and test sets to perform the attack. This simulates a real-world scenario where users of a private dataset train on the released dataset, and an attacker attempts to infer original samples from the private dataset. A lower attack success rate indicates a more robust generated dataset, meaning it performs better in protecting privacy against MIA.

\subsubsection{Utility metrics.}

As mentioned, we constructed a classification problem as the downstream task. To simulate a scenario where the data publisher is unaware of downstream tasks, we further evaluated the diversity and quality of the generated dataset, enabling adaptation to more complex tasks, such as data generation. We use a ResNet-18-based classifier to predict the class of the obfuscated image. Based on the channels and classes of the input dataset, we modified and replaced the input convolutional layer and the final fully connected layer. \textbf{Test accuracy} on the test dataset indicates the effectiveness of our obfuscation method in preserving discriminative features essential for classification. \textbf{Fréchet Inception Distance (FID)}\cite{chen2020gs} computes the Fréchet distance between original and generated image distributions in feature space. A lower FID indicates better quality and higher similarity. To evaluate the quality and diversity of the generated images, \textbf{Inception Score (IS)}\cite{salimans2016improved} classifies the generated images by pre-training the classifier, then calculates the entropy of the predicted class probability distribution for each image and the entropy of the average class probability distribution for all images. A higher IS denotes greater diversity. For both FID and IS scores, we used a pre-trained Inception V3 network for feature extraction. To accommodate the eight grayscale datasets, we expand each image to three channels and resized all datasets to the required dimensions (299x299) before conducting FID and IS evaluations. Additionally, we apply standard normalization using the ImageNet mean and standard deviation values to ensure compatibility with the pre-trained network's input requirements.

\subsection{Comparison with other Privacy-preserving Methods} 

This subsection presents the experimental results. Table \ref{tab:performance_metrics} provides the MIA success rates for downstream classification tasks, along with the test accuracy, FID scores, and IS scores for both the original dataset and the generated datasets processed by six different methods. Additionally, Fig. \ref{fig:comparison_samples_split} provides a visual quality comparison of the output samples generated by different methods. In the following three subsubsections, we systematically evaluate the comparative performance of our model against other privacy-preserving methods across three dimensions: privacy protection, data utility, and their inherent trade-off. The privacy analysis focuses on resistance to MIA attacks; the utility analysis covers image quality, diversity, and the classification accuracy of downstream models trained on the released dataset; the privacy-utility trade-off analysis is evaluated in conjunction with overall performance.

\renewcommand{\arraystretch}{0.82} 
\begin{longtable}{lcccc}
\caption{Performance Metrics for Different Models and Datasets} \label{tab:performance_metrics} \\
\toprule
\textbf{Model} & \textbf{MIA Success Rate (↓)} & \textbf{Test Acc (↑)} & \textbf{FID Score (↓)} & \textbf{IS Score (↑)} \\
\midrule
\endfirsthead

\multicolumn{5}{c}{\textbf{(Continued) Performance Metrics for Different Models and Datasets}} \\
\toprule
\textbf{Model} & \textbf{MIA Success Rate (↓)} & \textbf{Test Acc (↑)} & \textbf{FID Score (↓)} & \textbf{IS Score (↑)} \\
\midrule
\endhead

\bottomrule
\endfoot

\multicolumn{5}{c}{(Table continued on next page)} \\
\endfoot

\bottomrule
\endlastfoot
\multicolumn{5}{l}{\textbf{MNIST}} \\
Original     & 70.51\% & \textbf{97.85\%} & / & \textbf{2.5319} \\
Pixelation   & 58.98\% &  \textbf{97.12\%} & 1259.7742 & 1.5675 \\
Blur         & 65.63\%  & 94.13\% & 362.6897 & 2.2335 \\
K-anonymity  & \textbf{54.69\%} & \textbf{98.48\%} & 292.3094 & 2.2688 \\
VAEGAN-DP    & 66.56\% & 76.74\% & 451.0217 & \textbf{2.5719} \\
DPDM         & \textbf{55.21\%} & 96.62\% & \textbf{248.4533}  &  2.3321 \\
Ours         & \textbf{52.95\%} & 96.57\% & \textbf{116.2198} & \textbf{2.4618} \\
\hline
\multicolumn{5}{l}{\textbf{KMNIST}} \\
Original     & 67.73\% & \textbf{94.96\%} & / & \textbf{2.4417} \\
Pixelation   & 62.98\% & 90.89\% & 1241.3593 & 1.7498 \\
Blur         & 61.37\% & \textbf{93.80\%} & 420.9563 & \textbf{2.3659} \\
K-anonymity  & \textbf{56.60\%} & 91.14\% & 312.9801 & 2.0537 \\
VAEGAN-DP    & 63.98\% & 79.81\% & 662.5486 & \textbf{2.4978} \\
DPDM         & \textbf{61.28\%} & 90.64\% & \textbf{299.1109} & 2.2643 \\
Ours         & \textbf{56.59\%} & \textbf{93.22\%} & \textbf{148.8800} & 2.2833 \\
\hline
\multicolumn{5}{l}{\textbf{FMNIST}} \\
Original     & 65.63\% & \textbf{88.29\%} & / & \textbf{4.6867} \\
Pixelation   & 59.72\% & 86.27\% & 939.7923 & 1.7141 \\
Blur         & \textbf{56.25\%} &  \textbf{87.79\%} & 621.4550 & 2.6904 \\
K-anonymity  & \textbf{53.47\%} & 84.56\% & 482.9207 & 2.6112 \\
VAEGAN-DP    & 65.17\% & 65.79\% & 789.7272 & 2.8319 \\
DPDM         & 60.24\% & 80.69\% & \textbf{444.6866} &  \textbf{3.0555} \\
Ours         & \textbf{54.69\%} &  \textbf{87.37\%} & \textbf{234.6177} &  \textbf{3.3117} \\
\hline
\multicolumn{5}{l}{\textbf{notMNIST}} \\
Original     & 54.34\% & \textbf{95.62\%} & / & \textbf{3.9396} \\
Pixelation   & 51.22\% &  \textbf{94.66\%} & 1017.024 & 2.1068 \\
Blur         & \textbf{49.65\%} & \textbf{93.57\%} & 393.9886 & 3.0679 \\
K-anonymity  & 51.04\% & 86.50\% & \textbf{188.2756} & \textbf{3.7985} \\
VAEGAN-DP    & \textbf{50.87\%} & 78.80\% & 1407.0167 & 2.7642 \\
DPDM         & 51.91\% & 87.92\% & 1693.1748 & 2.6338  \\
Ours         & \textbf{49.65\%}&  89.37\% & \textbf{208.5805} &  \textbf{3.7380} \\
\hline
\multicolumn{5}{l}{\textbf{Math Symbols}} \\
Original     & 56.51\% & \textbf{99.74\%} & / & \textbf{2.8039} \\
Pixelation   & 54.08\% & \textbf{98.41\%} & 1309.5963 & 1.9217 \\
Blur         & 54.95\% & 82.35\% &  \textbf{379.6156} &  2.5815 \\
K-anonymity  & 54.60\% & 75.63\% & 379.6761 & 2.5685 \\
VAEGAN-DP    & \textbf{52.08\%} & 70.20\% & 1040.022 & 2.6180 \\
DPDM         & \textbf{52.08\%} & \textbf{92.56\%} & 424.9383 & \textbf{2.7880} \\
Ours         & \textbf{52.86\%}&  87.40\% & \textbf{293.0328} & \textbf{2.8568} \\
\hline
\multicolumn{5}{l}{\textbf{KANMNIST}} \\
Original     & 58.42\% & \textbf{98.59\%} & / & \textbf{2.6363} \\
Pixelation   & \textbf{55.47\%} & 91.37\% & 1068.388 & 1.7095 \\
Blur         & 56.34\%  &  93.96\% &  \textbf{272.1793} & 2.1170 \\
K-anonymity  & \textbf{54.42\%} & 82.93\% & 405.6101 & 2.0404 \\
VAEGAN-DP    & 57.73\% & 81.62\% & 362.4156 &  \textbf{2.4155} \\
DPDM         & 58.85\% & \textbf{94.92\%}  & 277.1584 & 1.9167 \\
Ours         & \textbf{55.64\%} & \textbf{94.27\%} & \textbf{227.5984} &  \textbf{2.2370} \\
\hline
\multicolumn{5}{l}{\textbf{AfroMNIST}} \\
Original     & 55.90\% & \textbf{99.96\%} & / & \textbf{1.7165} \\
Pixelation   & \textbf{50.52\%} & 71.64\% & 1340.0431 & 1.2620 \\
Blur         & \textbf{50.43\%} & 22.70\% & 527.1374 & 1.5693 \\
K-anonymity  & 50.95\%  & 79.16\% & 191.6329 & 1.6976 \\
VAEGAN-DP    & \textbf{49.31\%} & \textbf{87.74\%} & \textbf{124.7246} & \textbf{1.7157} \\
DPDM         & 51.39\% &  75.68\% &  \textbf{129.8099} & 1.4014 \\
Ours         & 52.52\% & \textbf{88.95\%} & 160.8828 & \textbf{1.7725} \\
\hline
\multicolumn{5}{l}{\textbf{dSprites}} \\
Original     & 52.08\% & \textbf{99.89\%} & / & \textbf{1.6985} \\
Pixelation   & \textbf{48.35\%} & \textbf{98.33\%} & 420.2735 & \textbf{1.5974} \\
Blur         & \textbf{49.13\%} & 69.40\% & 412.8178 & 1.3537 \\
K-anonymity  & 50.17\% & 73.62\% & 347.6233 & 1.3051 \\
VAEGAN-DP    & \textbf{49.65\%} & 78.25\% & 382.0260 & 1.2136 \\
DPDM         & 51.74\% & 99.18\% & \textbf{124.1060}  & 1.4568  \\
Ours         & 50.87\% & \textbf{99.73\%} & \textbf{65.9450} & \textbf{1.6300} \\
\hline
\multicolumn{5}{l}{\textbf{OCTMNIST}} \\
Original     & 64.75\% & \textbf{71.10\%} & / & \textbf{2.3262} \\
Pixelation   &  \textbf{59.72\%} & 47.80\% & 1025.7683 & 1.3694 \\
Blur         &  \textbf{63.37\%} & 34.10\% & 627.9113 & 1.4763 \\
K-anonymity  &  65.80\%  & 29.10\% & 548.8842 & 1.6482 \\
VAEGAN-DP    &  68.40\% & 32.00\% & 869.2015 & \textbf{1.9823} \\
DPDM         &  64.93\% & \textbf{49.00\%} & \textbf{113.3422} & 1.8090 \\
Ours         &  \textbf{52.26\%} & \textbf{56.50\%} & \textbf{361.8458} & \textbf{1.8595} \\
\hline
\multicolumn{5}{l}{\textbf{Average}} \\
Original     & 60.65\% & \textbf{94.00\%} & / & \textbf{2.7535} \\
Pixelation   & \textbf{55.67\%} & \textbf{86.28\%} & 1069.1132 & 1.6665 \\
Blur         & 56.35\% & 74.64\% & 446.5279 & 2.1617 \\
K-anonymity  & \textbf{54.64\%} & 77.90\% & \textbf{349.9903} & 2.2213 \\
VAEGAN-DP    & 58.19\% & 72.33\% & 676.5227 & \textbf{2.2901} \\
DPDM         & 56.40\% & 85.25\% & 417.1978 & 2.1842 \\
Ours         & \textbf{53.11\%} & \textbf{88.15\%} & \textbf{201.9559} & \textbf{2.4612} \\
\hline
\end{longtable}

\subsubsection{Privacy}

Table~\ref{tab:performance_metrics} presents MIA success rates across datasets and privacy protection methods. The original datasets have the highest attack success rates, averaging 60.65\%, highlighting severe privacy risks when no privacy protection techniques are applied.

Traditional obfuscation methods, such as pixelation and blurring, provide partial privacy protection. Pixelation effectively reduces the MIA success rate to an average of 55.67\%. However, it retains global structures, making it less effective for datasets like Math Symbols, where finer detail protection is necessary. Blurring performs well on notMNIST, reducing the attack success rate around 50\%, but is ineffective for datasets with larger feature scales, such as MNIST.

K-anonymity further lowers MIA success rates, averaging 54.64\%, with FMNIST achieving the best results at approximately 53\%. By replacing individual samples with cluster centers, it removes unique characteristics that aid membership inference. However, its effectiveness depends on parameter choices, such as the value of $K$ and the number of cluster centers. In datasets with high intra-class variance, such as OCTMNIST, protection is weaker, as indicated by an attack success rate of 65.80\%.

DP generative models yield mixed results. VAEGAN-DP achieves strong privacy on AfroMNIST but fails on OCTMNIST, where excessive noise hinders feature retention, leading to an attack success rate above 68\%. DPDM slightly improves privacy, reaching an average MIA success rate of 56.40\%, but remains less effective on OCTMNIST. While DP mechanisms introduce calibrated noise to enhance privacy, the final achieved privacy budget in our experiments is relatively high, resulting in less stringent privacy guarantees. Additionally, in datasets with large-scale identifiable features, DP-generated samples may retain distinguishing characteristics, limiting MIA protection.

Our method provides the most stable privacy protection, achieving the lowest average MIA success rate of 53.11\%. It performs especially well on OCTMNIST, significantly reducing attack success rates. This is attributed to structured perturbations that leverage curvature information to guide samples toward RBF centers, ensuring controlled movement away from high-curvature regions, thereby reducing identification risks.

\subsubsection{Utility}

\begin{figure*}
    \centering
    \begin{minipage}{0.49\textwidth}
        \centering
        \setlength{\tabcolsep}{0.8pt} 
        \renewcommand{\arraystretch}{0.8} 
        \footnotesize
        \begin{tabular}{>{\centering\arraybackslash}m{0.88cm} >{\centering\arraybackslash}m{0.8cm} >{\centering\arraybackslash}m{0.8cm} >{\centering\arraybackslash}m{0.8cm} >{\centering\arraybackslash}m{0.8cm} >{\centering\arraybackslash}m{0.8cm} >{\centering\arraybackslash}m{0.8cm} >{\centering\arraybackslash}m{0.8cm}}
            & Original & Ours & Pixel-ation & Blur & \mbox{K-anony} & VAEGAN-DP & DPDM \\

            MNIST &
            \includegraphics[width=0.8cm]{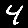} & 
            \includegraphics[width=0.8cm]{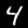} & 
            \includegraphics[width=0.8cm]{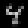} & 
            \includegraphics[width=0.8cm]{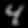} & 
            \includegraphics[width=0.8cm]{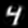} & 
            \includegraphics[width=0.8cm]{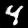} &
            \includegraphics[width=0.8cm]{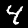}\\

            & \includegraphics[width=0.8cm]{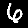} & 
            \includegraphics[width=0.8cm]{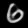} & 
            \includegraphics[width=0.8cm]{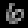} & 
            \includegraphics[width=0.8cm]{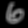} & 
            \includegraphics[width=0.8cm]{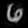} & 
            \includegraphics[width=0.8cm]{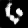} &
            \includegraphics[width=0.8cm]{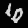}\\
            
            & \includegraphics[width=0.8cm]{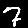} & 
            \includegraphics[width=0.8cm]{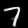} & 
            \includegraphics[width=0.8cm]{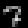} & 
            \includegraphics[width=0.8cm]{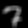} & 
            \includegraphics[width=0.8cm]{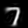} & 
            \includegraphics[width=0.8cm]{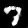} &
            \includegraphics[width=0.8cm]{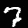}\\

            KMNIST &
            \includegraphics[width=0.8cm]{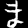} & 
            \includegraphics[width=0.8cm]{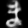} & 
            \includegraphics[width=0.8cm]{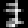} & 
            \includegraphics[width=0.8cm]{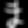} & 
            \includegraphics[width=0.8cm]{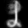} & 
            \includegraphics[width=0.8cm]{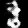} & 
            \includegraphics[width=0.8cm]{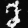}\\

            & \includegraphics[width=0.8cm]{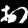} & 
            \includegraphics[width=0.8cm]{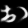} & 
            \includegraphics[width=0.8cm]{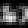} & 
            \includegraphics[width=0.8cm]{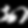} & 
            \includegraphics[width=0.8cm]{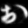} & 
            \includegraphics[width=0.8cm]{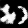} & 
            \includegraphics[width=0.8cm]{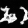}\\
            
            & \includegraphics[width=0.8cm]{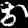} & 
            \includegraphics[width=0.8cm]{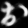} & 
            \includegraphics[width=0.8cm]{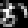} & 
            \includegraphics[width=0.8cm]{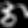} & 
            \includegraphics[width=0.8cm]{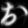} & 
            \includegraphics[width=0.8cm]{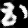} & 
            \includegraphics[width=0.8cm]{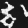}\\

            FMNIST &
            \includegraphics[width=0.8cm]{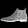} & 
            \includegraphics[width=0.8cm]{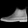} & 
            \includegraphics[width=0.8cm]{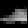} & 
            \includegraphics[width=0.8cm]{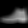} & 
            \includegraphics[width=0.8cm]{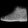} & 
            \includegraphics[width=0.8cm]{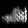} & 
            \includegraphics[width=0.8cm]{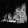} \\

            & \includegraphics[width=0.8cm]{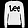} & 
            \includegraphics[width=0.8cm]{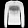} & 
            \includegraphics[width=0.8cm]{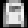} & 
            \includegraphics[width=0.8cm]{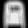} & 
            \includegraphics[width=0.8cm]{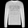} & 
            \includegraphics[width=0.8cm]{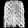} & 
            \includegraphics[width=0.8cm]{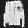} \\

            & \includegraphics[width=0.8cm]{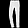} & 
            \includegraphics[width=0.8cm]{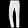} & 
            \includegraphics[width=0.8cm]{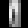} & 
            \includegraphics[width=0.8cm]{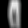} & 
            \includegraphics[width=0.8cm]{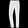} & 
            \includegraphics[width=0.8cm]{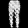} & 
            \includegraphics[width=0.8cm]{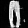} \\

            not-MNIST &
            \includegraphics[width=0.8cm]{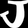} & 
            \includegraphics[width=0.8cm]{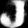} & 
            \includegraphics[width=0.8cm]{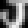} & 
            \includegraphics[width=0.8cm]{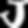} & 
            \includegraphics[width=0.8cm]{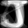} & 
            \includegraphics[width=0.8cm]{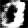} & 
            \includegraphics[width=0.8cm]{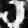} \\

            & \includegraphics[width=0.8cm]{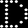} & 
            \includegraphics[width=0.8cm]{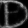} & 
            \includegraphics[width=0.8cm]{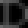} & 
            \includegraphics[width=0.8cm]{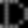} & 
            \includegraphics[width=0.8cm]{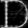} & 
            \includegraphics[width=0.8cm]{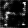} & 
            \includegraphics[width=0.8cm]{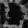} \\
            
            & \includegraphics[width=0.8cm]{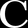} & 
            \includegraphics[width=0.8cm]{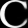} & 
            \includegraphics[width=0.8cm]{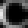} & 
            \includegraphics[width=0.8cm]{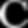} & 
            \includegraphics[width=0.8cm]{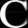} & 
            \includegraphics[width=0.8cm]{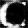} & 
            \includegraphics[width=0.8cm]{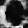} \\

            Math Symbols &
        \includegraphics[width=0.8cm]{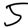} & 
        \includegraphics[width=0.8cm]{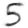} & 
        \includegraphics[width=0.8cm]{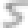} & 
        \includegraphics[width=0.8cm]{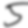} & 
        \includegraphics[width=0.8cm]{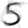} & 
        \includegraphics[width=0.8cm]{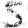} & 
        \includegraphics[width=0.8cm]{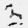} \\

        & \includegraphics[width=0.8cm]{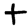} & 
        \includegraphics[width=0.8cm]{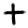} & 
        \includegraphics[width=0.8cm]{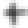} & 
        \includegraphics[width=0.8cm]{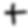} & 
        \includegraphics[width=0.8cm]{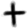} & 
        \includegraphics[width=0.8cm]{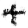}  & 
        \includegraphics[width=0.8cm]{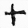} \\
        
        & \includegraphics[width=0.8cm]{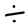} & 
        \includegraphics[width=0.8cm]{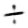} & 
        \includegraphics[width=0.8cm]{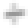} & 
        \includegraphics[width=0.8cm]{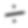} & 
        \includegraphics[width=0.8cm]{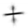} & 
        \includegraphics[width=0.8cm]{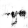}  & 
        \includegraphics[width=0.8cm]{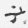} \\

        \end{tabular}
    \end{minipage}
        \begin{minipage}{0.49\textwidth}
        \centering
        \setlength{\tabcolsep}{0.8pt} 
        \renewcommand{\arraystretch}{0.8} 
        \footnotesize
        \begin{tabular}{>{\centering\arraybackslash}m{0.8cm} >{\centering\arraybackslash}m{0.8cm} >{\centering\arraybackslash}m{0.8cm} >{\centering\arraybackslash}m{0.8cm} >{\centering\arraybackslash}m{0.8cm} >{\centering\arraybackslash}m{0.8cm} >{\centering\arraybackslash}m{0.8cm} >{\centering\arraybackslash}m{0.8cm}}
            & Original & Ours & Pixel-ation & Blur & \mbox{K-anony} & VAEGAN-DP & DPDM \\
            KAN-MNIST &
            \includegraphics[width=0.8cm]{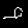} & 
            \includegraphics[width=0.8cm]{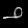} & 
            \includegraphics[width=0.8cm]{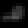} & 
            \includegraphics[width=0.8cm]{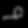} & 
            \includegraphics[width=0.8cm]{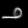} & 
            \includegraphics[width=0.8cm]{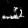}  & 
            \includegraphics[width=0.8cm]{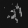} \\

            & \includegraphics[width=0.8cm]{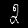} & 
            \includegraphics[width=0.8cm]{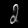} & 
            \includegraphics[width=0.8cm]{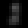} & 
            \includegraphics[width=0.8cm]{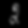} & 
            \includegraphics[width=0.8cm]{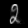} & 
            \includegraphics[width=0.8cm]{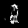} & 
            \includegraphics[width=0.8cm]{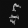} \\
            
            & \includegraphics[width=0.8cm]{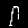} & 
            \includegraphics[width=0.8cm]{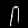} & 
            \includegraphics[width=0.8cm]{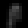} & 
            \includegraphics[width=0.8cm]{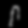} & 
            \includegraphics[width=0.8cm]{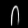} & 
            \includegraphics[width=0.8cm]{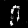} & 
            \includegraphics[width=0.8cm]{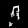} \\

            Afro-MNIST &
            \includegraphics[width=0.8cm]{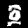} & 
            \includegraphics[width=0.8cm]{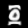} & 
            \includegraphics[width=0.8cm]{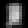} & 
            \includegraphics[width=0.8cm]{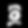} & 
            \includegraphics[width=0.8cm]{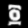} & 
            \includegraphics[width=0.8cm]{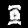} & 
            \includegraphics[width=0.8cm]{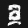} \\

            & \includegraphics[width=0.8cm]{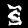} & 
            \includegraphics[width=0.8cm]{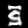} & 
            \includegraphics[width=0.8cm]{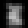} & 
            \includegraphics[width=0.8cm]{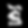} & 
            \includegraphics[width=0.8cm]{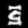} & 
            \includegraphics[width=0.8cm]{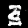} & 
            \includegraphics[width=0.8cm]{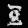} \\
            
            & \includegraphics[width=0.8cm]{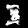} & 
            \includegraphics[width=0.8cm]{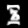} & 
            \includegraphics[width=0.8cm]{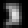} & 
            \includegraphics[width=0.8cm]{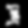} & 
            \includegraphics[width=0.8cm]{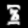} & 
            \includegraphics[width=0.8cm]{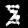} & 
            \includegraphics[width=0.8cm]{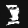} \\

            dSprites &
            \includegraphics[width=0.8cm]{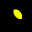} & 
            \includegraphics[width=0.8cm]{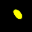} & 
            \includegraphics[width=0.8cm]{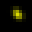} & 
            \includegraphics[width=0.8cm]{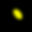} & 
            \includegraphics[width=0.8cm]{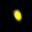} & 
            \includegraphics[width=0.8cm]{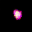} & 
            \includegraphics[width=0.8cm]{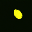} \\

            & \includegraphics[width=0.8cm]{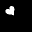} & 
            \includegraphics[width=0.8cm]{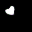} & 
            \includegraphics[width=0.8cm]{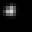} & 
            \includegraphics[width=0.8cm]{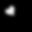} & 
            \includegraphics[width=0.8cm]{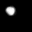} & 
            \includegraphics[width=0.8cm]{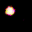} & 
            \includegraphics[width=0.8cm]{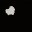} \\
            
            & \includegraphics[width=0.8cm]{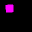} & 
            \includegraphics[width=0.8cm]{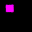} & 
            \includegraphics[width=0.8cm]{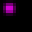} & 
            \includegraphics[width=0.8cm]{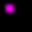} & 
            \includegraphics[width=0.8cm]{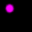} & 
            \includegraphics[width=0.8cm]{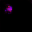} & 
            \includegraphics[width=0.8cm]{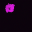} \\

            OCT-MNIST &
            \includegraphics[width=0.8cm]{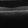} & 
            \includegraphics[width=0.8cm]{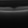} & 
            \includegraphics[width=0.8cm]{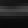} & 
            \includegraphics[width=0.8cm]{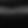} & 
            \includegraphics[width=0.8cm]{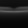} & 
            \includegraphics[width=0.8cm]{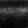} & 
            \includegraphics[width=0.8cm]{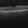} \\

            & \includegraphics[width=0.8cm]{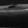} & 
            \includegraphics[width=0.8cm]{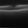} & 
            \includegraphics[width=0.8cm]{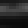} & 
            \includegraphics[width=0.8cm]{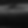} & 
            \includegraphics[width=0.8cm]{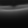} & 
            \includegraphics[width=0.8cm]{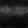} & 
            \includegraphics[width=0.8cm]{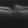} \\
            
            & \includegraphics[width=0.8cm]{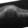} & 
            \includegraphics[width=0.8cm]{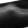} & 
            \includegraphics[width=0.8cm]{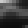} & 
            \includegraphics[width=0.8cm]{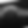} & 
            \includegraphics[width=0.8cm]{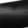} & 
            \includegraphics[width=0.8cm]{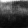} & 
            \includegraphics[width=0.8cm]{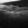} \\

            & \includegraphics[width=0.8cm]{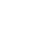} & 
            \includegraphics[width=0.8cm]{blank/blank.png} & 
            \includegraphics[width=0.8cm]{blank/blank.png}& 
            \includegraphics[width=0.8cm]{blank/blank.png} & 
            \includegraphics[width=0.8cm]{blank/blank.png} & 
            \includegraphics[width=0.8cm]{blank/blank.png} & 
            \includegraphics[width=0.8cm]{blank/blank.png} \\

            & \includegraphics[width=0.8cm]{blank/blank.png} & 
            \includegraphics[width=0.8cm]{blank/blank.png} & 
            \includegraphics[width=0.8cm]{blank/blank.png}& 
            \includegraphics[width=0.8cm]{blank/blank.png} & 
            \includegraphics[width=0.8cm]{blank/blank.png} & 
            \includegraphics[width=0.8cm]{blank/blank.png} & 
            \includegraphics[width=0.8cm]{blank/blank.png} \\

            & \includegraphics[width=0.8cm]{blank/blank.png} & 
            \includegraphics[width=0.8cm]{blank/blank.png} & 
            \includegraphics[width=0.8cm]{blank/blank.png}& 
            \includegraphics[width=0.8cm]{blank/blank.png} & 
            \includegraphics[width=0.8cm]{blank/blank.png} & 
            \includegraphics[width=0.8cm]{blank/blank.png} & 
            \includegraphics[width=0.8cm]{blank/blank.png} \\
            
        \end{tabular}
    \end{minipage}
    \caption{Comparison of samples for different models and datasets(Left: MNIST, KMNIST, FMNIST, notMNIST, Math Symbols. Right: KAN-MNIST, Afro-MNIST, dSprites, OCTMNIST)}
    \label{fig:comparison_samples_split}
\end{figure*}

Table~\ref{tab:performance_metrics} compares classification accuracy, FID, and IS scores across methods. Fig.~\ref{fig:comparison_samples_split} presents a qualitative comparison of output samples. The table indicates that the original datasets have the highest utility, demonstrating superior performance in accuracy, FID, and IS scores, with AfroMNIST achieving near-perfect classification.

Pixelation and blurring preserve classification accuracy but degrade sample quality. Pixelation achieves over 97\% accuracy on MNIST but produces coarse images, leading to an extremely high FID. Fig.~\ref{fig:comparison_samples_split} shows that pixelation makes MNIST digits blocky, reducing detail clarity. Blurring retains structural features but causes indistinct shapes, particularly in KANMNIST, where character edges become unclear.

K-anonymity can generate samples with more realistic details, as reflected in its lower FID score. However, it relies on replacing original samples with cluster centers, which sacrifices sample diversity. When the dataset is small and exhibits high intra-cluster variance, the decline in utility becomes more pronounced. For example, in OCTMNIST, classification accuracy drops by 40\%.

VAEGAN-DP introduces noise, degrading both classification and sample quality. In KMNIST, its reconstructions appear grainy, obscuring fine details. It performs worst on OCTMNIST, where classification accuracy drops below 35\%. This degradation occurs because excessive noise reduces interpretability, ultimately impacting classification reliability. Benefiting from a more robust backbone, DPDM demonstrates superior utility performance compared to VAEGAN-DP. DPDM improves classification accuracy to an average of 85.25\% and reduces FID but still suffers from privacy-related noise. As shown in Fig.~\ref{fig:comparison_samples_split}, the first DPDM sample in the Math Symbols dataset becomes difficult to recognize due to noise.

Our method achieves the highest average test accuracy, the lowest FID, and the highest IS score among privacy-preserving models. Fig.~\ref{fig:comparison_samples_split} also illustrates that our method effectively preserves structural integrity while enhancing fidelity and diversity, making it the most effective approach among the evaluated methods.

\subsubsection{Privacy-Utility Trade-off} 

Different methods exhibit distinct trade-offs between privacy and utility, revealing inherent limitations in current approaches. The original dataset provides optimal utility but is also the most vulnerable to MIA attacks, underscoring the fundamental challenge of designing privacy-preserving mechanisms without sacrificing utility.

Traditional techniques such as pixelation and blurring represent simple generalization strategies, where the trade-off primarily manifests as a severe degradation in sample quality in exchange for partial utility retention. Pixelation achieves a relatively high classification accuracy of 86.28\%, but its excessively high FID values indicate a significant loss in sample quality. Additionally, these methods offer limited privacy protection, making them ineffective trade-off solutions.

K-anonymity attempts to achieve a more balanced trade-off between privacy and utility, reducing the MIA success rate to 54.64\% while maintaining a classification accuracy of 77.90\%. However, its performance is highly sensitive to dataset characteristics and parameter configurations, leading to inconsistencies, especially when applied to complex distributions such as OCTMNIST.

DP-based generative models, on the other hand, exemplify the extreme end of the trade-off spectrum by prioritizing privacy at the cost of utility. VAEGAN-DP achieves only a 2\% reduction in MIA success rate while significantly compromising utility, with classification accuracy dropping to 72.33\% and reconstructions exhibiting excessive noise. In contrast, DPDM benefits from a more powerful generative backbone, improving utility (85.25\% accuracy with better FID scores), but its privacy protection remains inconsistent, as evidenced by the notably high MIA success rate in OCTMNIST.

In comparison, our method achieves the best trade-off, reducing the MIA success rate to 53.11\% while maintaining the highest classification accuracy of 88.15\%, the lowest FID, and the highest IS. This demonstrates superior optimization in simultaneously preserving privacy and retaining utility, even for datasets with extreme variations such as OCTMNIST. Fig.~\ref{fig:comparison_samples_split} further illustrates this in FMNIST, where our method effectively removes individual-specific features, such as T-shirt text, while preserving the overall structure and maintaining edge clarity, such as the sleeves.

\subsection{Visualization of Latent Space and Geodesics}

\begin{figure*} [!htbp]
    \centering
    \subfigure[2D latent space]{%
        \begin{minipage}{0.51\textwidth}
            \centering
            \includegraphics[width=\textwidth]{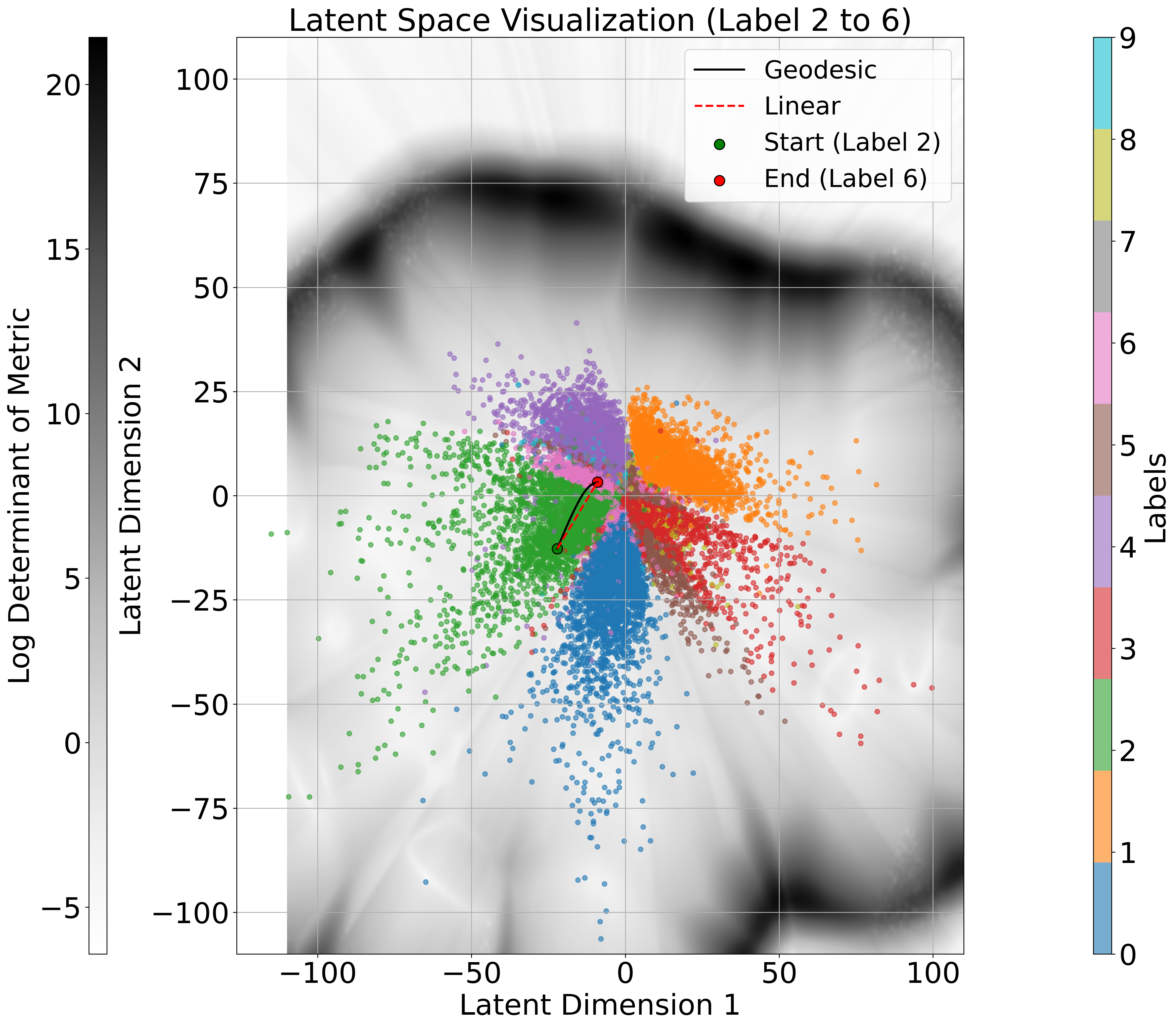}
        \end{minipage}%
    }
    \hfill
    \subfigure[Linear interpolation samples \& geodesic samples]{%
        \begin{minipage}{0.48\textwidth}
            \centering
            \raisebox{0.3\height}{\includegraphics[width=\textwidth]{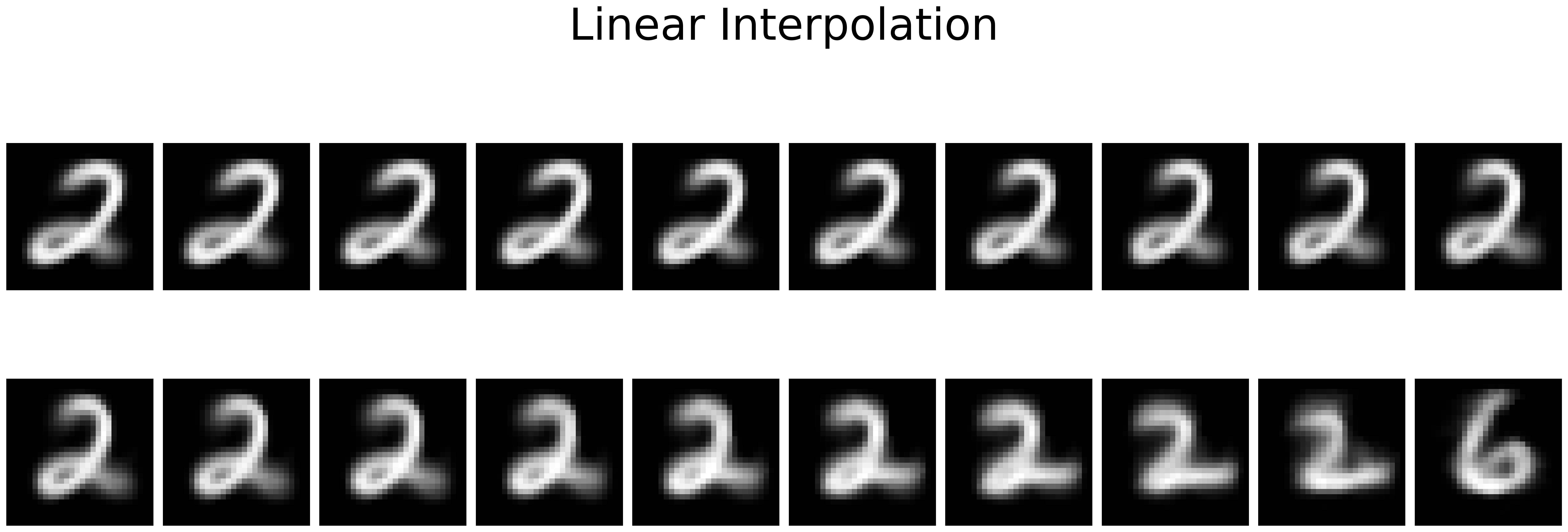}} \\
            \includegraphics[width=\textwidth]{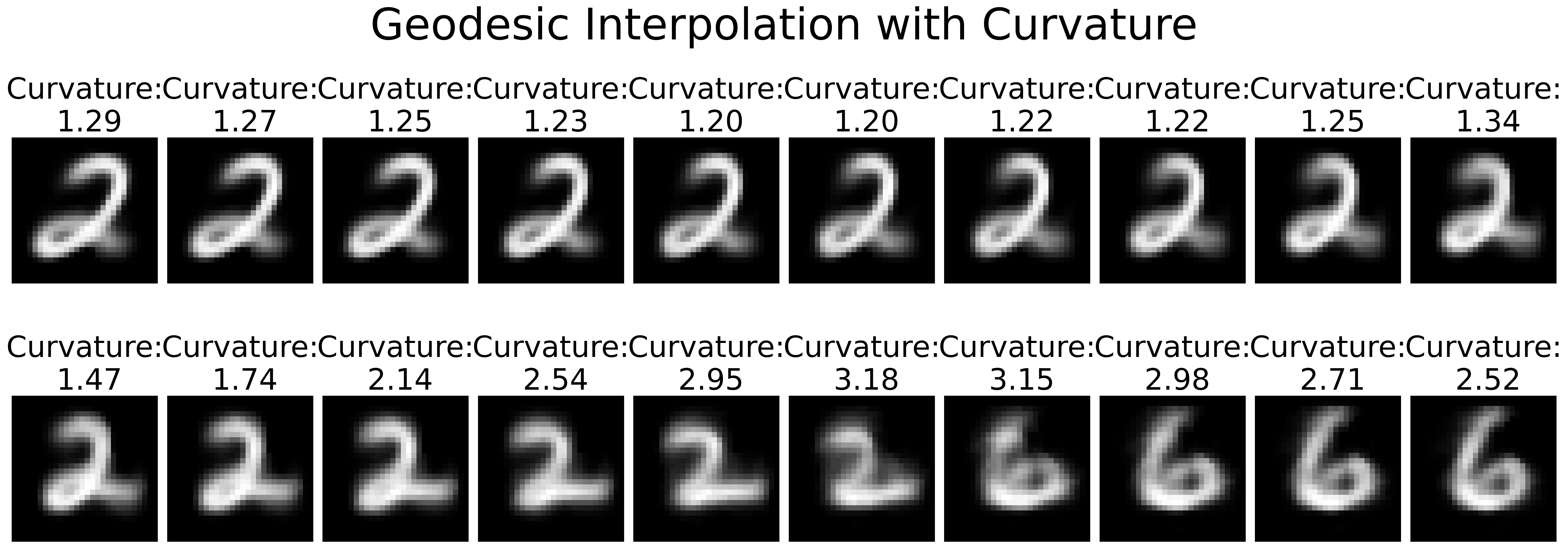}
        \end{minipage}%
    }
    
    \caption{Latent space and geodesic visualization (latent dim=2).}
    \label{fig:2D_latent}
\end{figure*}

To provide a clearer understanding of the operation of our geodesic obfuscator, including the performance of the curvature estimator, and the effect of geodesic interpolation, we present a visualization of the latent space. As shown in Fig. \ref{fig:2D_latent}, to facilitate visualization in a 2D plane, we changed the latent variable dimension from 10 to 2 in the MNIST dataset experiment and visualized the geodesic and linear interpolations between samples labeled 2 and 6 to reveal the nature of the data movement in latent space. From Fig. \ref{fig:2D_latent}(a), it can be observed that the geodesic bends at the class boundary, corresponding to the second image in Fig. \ref{fig:2D_latent}(b). Here, the curvature increases at the class boundary, validating the hypothesis that high curvature may occur at decision boundaries. At the same time, Fig. \ref{fig:2D_latent}(b) also shows that the clustering within class 6 has higher intrinsic curvature. This is because we down-sampled the 6-9 classes, making class 6 part of the distribution tail. This aligns with our expectations of higher curvature.

\begin{figure}
    \centering
    \subfigure[Geodesic (latent dim=10, using ISOMAP for dimensionality reduction)]{%
        \begin{minipage}{0.54\linewidth}
            \centering
            \includegraphics[width=\linewidth]{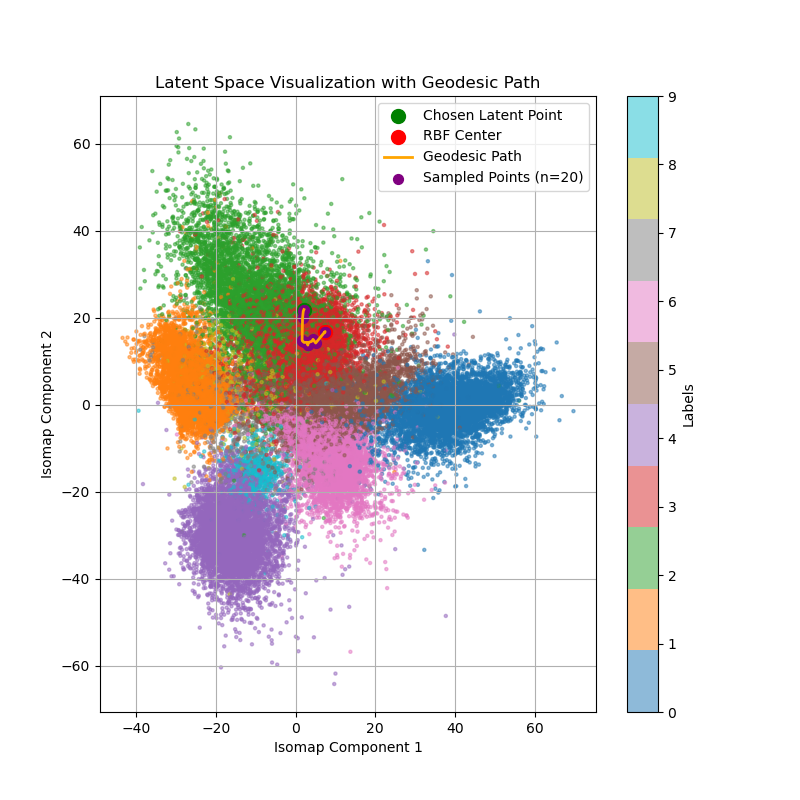}
        \end{minipage}%
    }
    \hfill 
    \subfigure[Samples along geodesic]{%
        \begin{minipage}{0.4\linewidth}
            \centering
            \includegraphics[width=\linewidth]{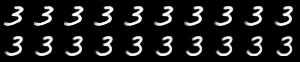}
        \end{minipage}%
    }
    
    \caption{Visualization of geodesic and sample distributions.}
    \label{fig:geodesic_side_by_side}
\end{figure}

The geodesic, shown in black in Fig. \ref{fig:2D_latent}(a), provides a smooth transition, preserving the underlying structure of the manifold and navigating seamlessly between different regions of the latent space. This indicates that the geodesic method is more effective at capturing local curvature and relationships between data points. The gradual and smooth transition from digit 2 to digit 6 demonstrates our model's ability to generate coherent representations using geodesics. By contrast, the linear interpolation, shown in red dashed line in Fig. \ref{fig:2D_latent}(a), while more direct, does not produce as smooth a transition between class features as the geodesic, as evidenced by the visualized samples.

As shown in the Fig. \ref{fig:geodesic_side_by_side} (a), this is the result of applying Isometric Mapping(ISOMAP) dimensionality reduction to our 10-dimensional latent space. We performed geodesic interpolation between the sample and its corresponding RBF center, which contributes the most to its local representation in the latent space. Moreover, typically, the selected RBF center with the highest contribution will also fall within the same class. The samples generated along the geodesic are displayed in Fig. \ref{fig:geodesic_side_by_side} (b). It is evident that our model aims to generate samples that remain within the class boundaries and exhibit more common, generalized features.

\subsection{Comparative Analysis of Interpolation Strategies}

\begin{table*}[!htbp]
    \centering
    \begin{minipage}{0.6\linewidth}
        \centering
        \caption{Performance Comparison for Different Interpolation Strategies for MNIST.}
        \label{tab:model_comparison}
        \renewcommand{\arraystretch}{1.2} 
        \setlength{\tabcolsep}{1pt} 
        \begin{tabular}{c c c c c}
            \hline
            \textbf{Model} & 
            \makecell{\textbf{MIA Success} \\ \textbf{Rate} (↓)} & 
            \makecell{\textbf{Test} \\ \textbf{Acc} (↑)} &  
            \makecell{\textbf{FID Score} \\ (↓)} & 
            \makecell{\textbf{IS Score} \\ (↑)} \\
            \hline
            Original   & 70.51\% & 97.85\% & /         & 2.5319 \\
            Linear     & 53.65\% & 62.08\% & 232.5755  & 2.3997 \\
            Highest curvature  & 58.68\% & 89.40\% & 178.1020  & 2.4985 \\
            Ours       & 52.95\% & 96.57\% & 116.2198  & 2.4618 \\
            \hline
        \end{tabular}
    \end{minipage}
\end{table*}

\begin{figure}[!htbp]
    \centering
    \setlength{\tabcolsep}{1pt} 
    \renewcommand{\arraystretch}{1} 

    \begin{tabular}{>{\centering\arraybackslash}m{1.5cm} >{\centering\arraybackslash}m{1.5cm} >{\centering\arraybackslash}m{1.5cm} >{\centering\arraybackslash}m{1.5cm} >{\centering\arraybackslash}m{1.5cm}}

        \textbf{Original} & \textbf{Ours} & \textbf{Highest Curvature} & \textbf{Linear} & \textbf{RBF Center} \\

        \includegraphics[width=1.4cm]{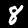} & 
        \includegraphics[width=1.4cm]{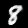} & 
        \includegraphics[width=1.4cm]{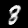} & 
        \includegraphics[width=1.4cm]{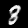} & 
        \includegraphics[width=1.4cm]{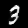} \\

        \includegraphics[width=1.4cm]{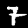} & 
        \includegraphics[width=1.4cm]{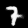} & 
        \includegraphics[width=1.4cm]{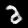} & 
        \includegraphics[width=1.4cm]{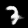} & 
        \includegraphics[width=1.4cm]{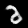} \\

        \includegraphics[width=1.4cm]{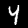} & 
        \includegraphics[width=1.4cm]{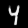} & 
        \includegraphics[width=1.4cm]{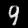} & 
        \includegraphics[width=1.4cm]{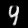} & 
        \includegraphics[width=1.4cm]{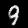} \\

    \end{tabular}

    \caption{Comparison of samples for different interpolation strategies for MNIST.}
    \label{fig:ablation_study}
\end{figure}



We further analyzed the perfomances of different interpolation strategies, as summarized in Table~\ref{tab:model_comparison}. The original, unperturbed data exhibited the highest classification accuracy of 97.85\% but suffered from the worst privacy vulnerability, with an MIA success rate of 70.51\%. Linear interpolation significantly reduced MIA success rate to 53.65\%, but at the cost of classification performance, which dropped to 62.08\%, and a higher FID score of 232.58, indicating poor sample quality. Interpolating towards the highest curvature regions improved sample fidelity, as reflected in a lower FID score of 178.10 and a classification accuracy of 89.40\%. This improvement is attributed to the precision of geodesic interpolation. The IS score increase may result from interpolating into high-curvature regions, generating diverse features distinct from the original domain. However, this method also increased the risk of successful MIA attacks, with the MIA success rate rising to 58.68\%, indicating that the generated samples are closer to decision boundaries, distinctive feature combinations, and other high-risk regions, making them more susceptible to attacks. In contrast, our method, which perturbs samples along geodesics towards the lowest curvature regions, achieved the lowest MIA success rate of 52.95\% while maintaining high classification accuracy at 96.57\%. Additionally, it produced the lowest FID score of 116.22, confirming that the generated samples preserve their structure and visual coherence better than other interpolation strategies.

Fig~\ref{fig:ablation_study} provides a qualitative comparison of sample outputs across different perturbation strategies. The original images retain their sharpness and details, whereas linear interpolation introduces significant distortions. Perturbation towards the highest curvature regions results in samples that are closer to the decision boundary, making them the least representative of actual numerical values. In contrast, our method preserves fine details while ensuring sufficient transformation for privacy protection.


\subsection{Analysis of Extrinsic Curvature and Sample Vulnerability}

\begin{figure}[!htbp]
    \centering
    \includegraphics[width=0.72\linewidth]{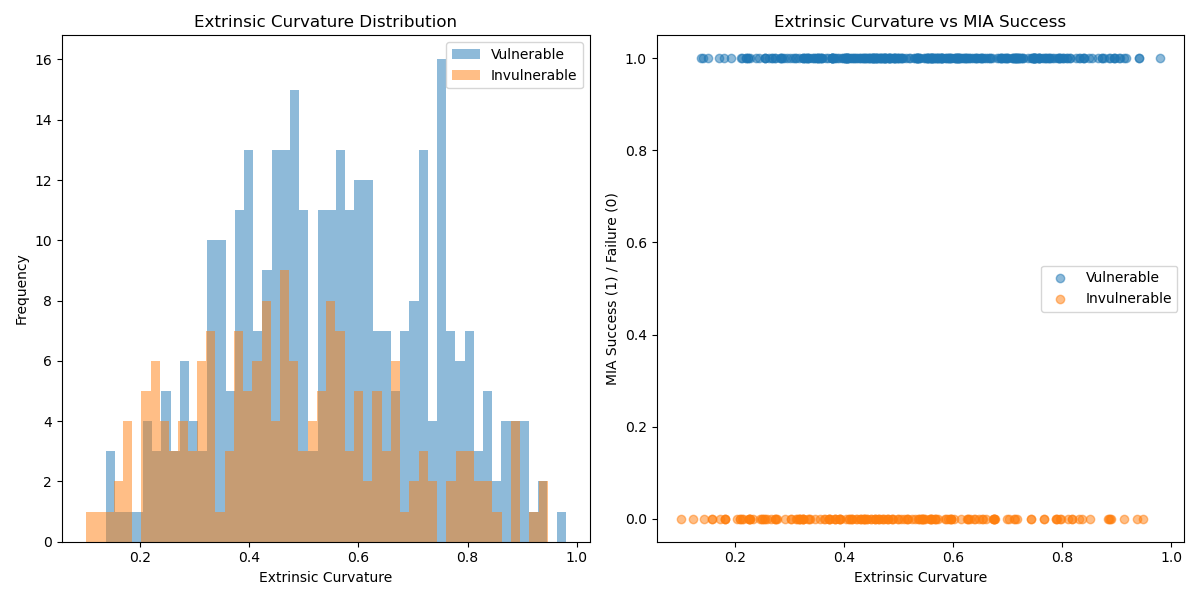}
    \caption{Distribution of extrinsic curvature for successful and failed MIA samples}
    \label{fig:curvature_distribution}
\end{figure}

\begin{figure}[!htbp]
    \centering
    \setlength{\tabcolsep}{1pt} 
    \renewcommand{\arraystretch}{1} 
    \begin{tabular}{>{\centering\arraybackslash}m{2.1cm} >{\centering\arraybackslash}m{1.5cm} >{\centering\arraybackslash}m{1.5cm} >{\centering\arraybackslash}m{1.5cm} >{\centering\arraybackslash}m{1.5cm} >{\centering\arraybackslash}m{1.5cm}}

        & \small{\textbf{Image 1}} & \small{\textbf{Image 2}} & \small{\textbf{Image 3}} & \small{\textbf{Image 4}} & \small{\textbf{Image 5}}\\

        \small{\textbf{Vulnerable Samples}} &
        \includegraphics[width=1.4cm]{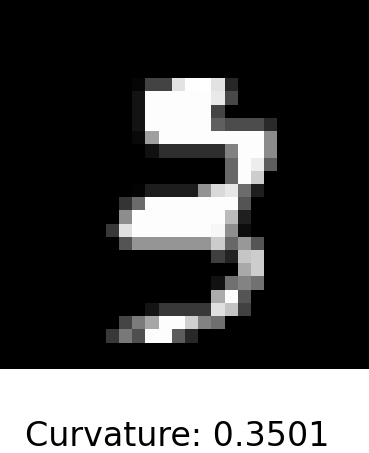} & 
        \includegraphics[width=1.4cm]{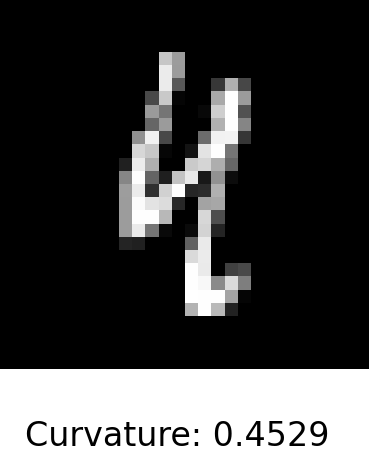}  &
        \includegraphics[width=1.4cm]{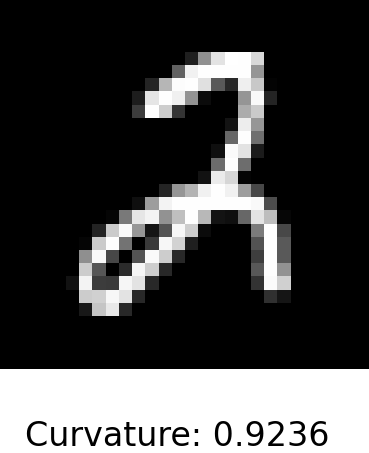} &
        \includegraphics[width=1.4cm]{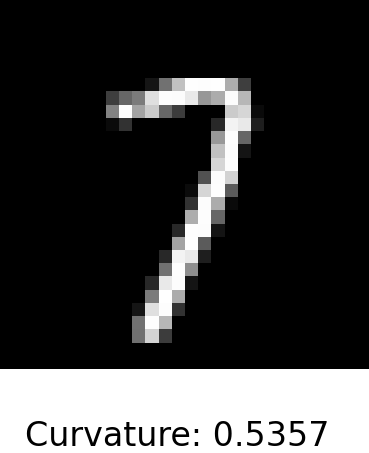} & 
        \includegraphics[width=1.4cm]{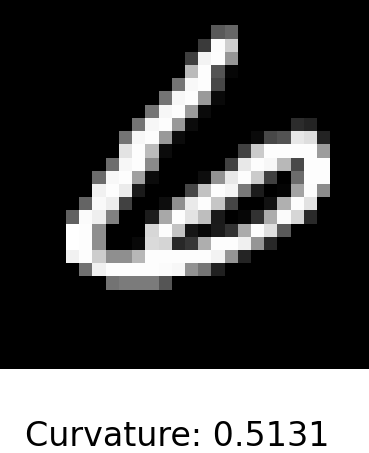} \\

        \small{\textbf{Invulnerable Samples}} &
        \includegraphics[width=1.4cm]{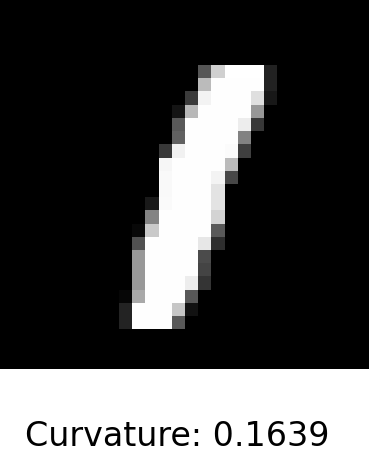} & 
        \includegraphics[width=1.4cm]{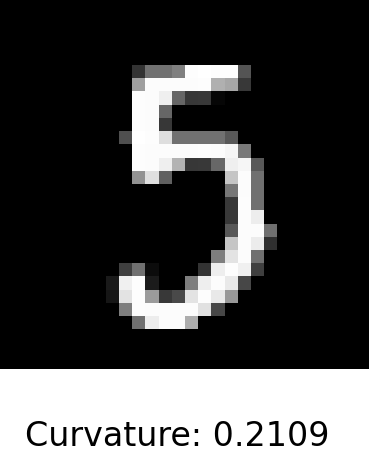} & 
        \includegraphics[width=1.4cm]{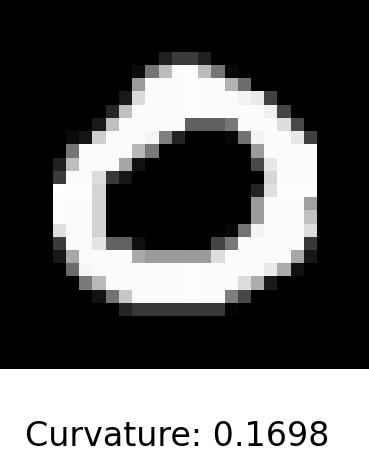} &
        \includegraphics[width=1.4cm]{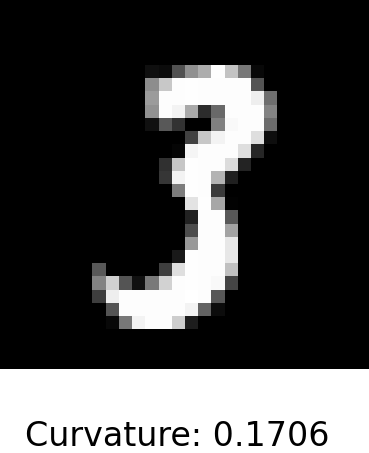}  &
        \includegraphics[width=1.4cm]{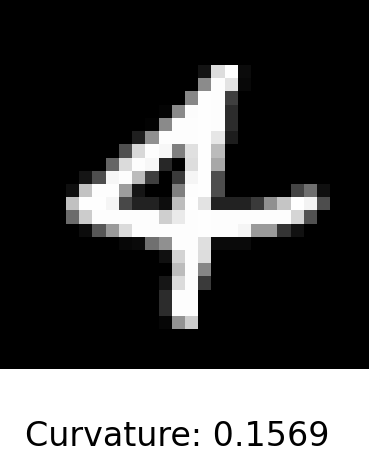}\\
        
    \end{tabular}

    \caption{Vulnerable and Invulnerable MNIST Samples}
    \label{fig:vulnerable_invulnerable_samples}
\end{figure}

In this subsection, we aim to analyze the relationship between extrinsic curvature of the data manifold and the vulnerable samples. We conducted an MIA evaluation on the original MNIST dataset, estimating the extrinsic curvature in the local region of each data point to capture the subtle geometric characteristics related to MIA susceptibility. We then approximated the estimates of the extrinsic curvature through ISOMAP dimensionality reduction and the eigenvalue ratio of the local covariance matrix. The average extrinsic curvature for vulnerable samples was 0.0629, and 0.0554 for the invulnerable samples, yielding a correlation of 0.1636. This indicates a certain level of association. The moderate correlation primarily results from the inherent randomness in loss-based MIA predictions, despite the observable curvature difference between vulnerable and invulnerable samples. Fig. \ref{fig:curvature_distribution} illustrates that the samples vulnerable to an MIA tend to have a slightly higher extrinsic curvature than the non-vulnerable samples. Fig. \ref{fig:vulnerable_invulnerable_samples} shows several different samples alongside their corresponding extrinsic curvature values. From this, we can see that samples with a higher extrinsic curvature often display more complex and unique handwriting styles or belong to minority classes in a long-tailed distribution (specifically, the down-sampled classes 6-9 in our processing).




\section{Conclusion}

In this study, we present a bilevel optimization framework for privacy-preserving data publishing, which effectively balances privacy protection and data utility. The upper-level task focuses on maintaining high data quality through discriminator-guided optimization, while the lower-level task targets privacy protection through curvature-guided geodesic obfuscation. For the lower-level task, our method leverages manifold learning within the RVAE framework, enabling geodesic interpolation in latent space while avoiding high-curvature regions associated with vulnerable data points. This strategy ensures precise perturbation control to mitigate privacy risks. Experimental results demonstrate that our framework outperforms existing methods in multiple aspects: it maintains robust resistance to MIA while preserving essential data features and sample diversity. The framework's ability to generate high-quality synthetic samples highlights its potential for various privacy-sensitive applications.

Although the RVAE provides robust manifold-capturing capabilities and high-quality generation, its scalability to high-resolution datasets is hindered by the computational burden of the Jacobian matrix. As a result, current research on RVAEs is largely confined to grayscale dataset generation. To overcome this limitation, a more computationally efficient Riemannian metric is required. Future work will explore alternative metrics to replace the pullback metric and expand our method’s applicability.

\section{Acknowledgments}

This work is supported by the Australian Research Council under Discovery Project DP230101540.

\bibliographystyle{ACM-Reference-Format}  
\bibliography{sample-base}  

\appendix
\section*{Appendix}
\section{Mathematical Analysis of High-Curvature Data Manifold Regions and MIA Vulnerability}
\label{proof}

In this section of the appendix, we demonstrate that samples located in high-curvature regions of the data manifold exhibit significantly larger loss variations even under infinitesimal perturbations. This characteristic makes their loss more distinguishable from those of non-member samples, thereby enabling attackers to more precisely identify these samples using loss-based MIA methods.

Assume the RVAE decoder is defined as:
\begin{equation}
    x = f_{\theta}(z),
\end{equation}
which maps the latent variable \( z \) to the data space \( x \), learning a low-dimensional manifold \( \mathcal{M} \). The downstream classifier is defined as \( y = g_{\phi}(x) \), and the loss function is given by:
\begin{equation}
    L(x) = L(g_{\phi}(x), y).
\end{equation}

Our objective is to demonstrate that when the decoder Hessian \( H_{\text{decoder}}(z) \) has large eigenvalues (indicating high-curvature regions), the variation in classification loss, denoted as \( \Delta L \), exhibits heightened sensitivity to perturbations in the latent space, meaning that larger curvature results in greater fluctuations in \( L(x) \).

For a perturbation in the data space, given by \( x' = x + \delta x \), the loss function can be expanded to the second order:
\begin{equation}
    L(x + \delta x) = L(x) + \nabla L(x)^T \delta x + \frac{1}{2} \delta x^T H_{\text{classifier}}(x) \delta x + O(\|\delta x\|^3),
\end{equation}
where \( H_{\text{classifier}}(x) = \frac{\partial^2 L}{\partial x^2} \) denotes the classifier Hessian. The variation in loss is then given by:
\begin{equation}
    \Delta L = L(x + \delta x) - L(x) = \nabla L(x)^T \delta x + \frac{1}{2} \delta x^T H_{\text{classifier}}(x) \delta x + O(\|\delta x\|^3).
\end{equation}
For training samples, when the training of classifier is completed, the gradient term typically vanishes, i.e., \( \nabla L(x) \approx 0 \), leading to:
\begin{equation}
    \Delta L = \frac{1}{2} \delta x^T H_{\text{classifier}}(x) \delta x + O(\|\delta x\|^3).
\end{equation}

The data perturbation \( \delta x \) arises from a latent perturbation \( \delta z \) through the decoder mapping. Expanding \( \delta x \) to the second order:
\begin{equation}
    \delta x = J_{\theta} \delta z + \frac{1}{2} \delta z^T H_{\text{decoder}}(z) \delta z + O(\|\delta z\|^3),
\end{equation}
where \( J_{\theta} = \frac{\partial f_{\theta}(z)}{\partial z} \) is the Jacobian matrix and \( H_{\text{decoder}}(z) = \frac{\partial^2 f_{\theta}(z)}{\partial z^2} \) is the decoder Hessian.

High-curvature regions correspond to large eigenvalues of \( H_{\text{decoder}}(z) \), enhancing local geometric nonlinearity. This implies that for the same latent perturbation \( \delta z \), the resulting data perturbation \( \delta x \) becomes more significant. Specifically:
\begin{enumerate}
    \item When \( H_{\text{decoder}}(z) \) is large, the second-order term \( \frac{1}{2} \delta z^T H_{\text{decoder}}(z) \delta z \) dominates the perturbation in data space.
    \item Since \( \Delta L \propto \delta x^T H_{\text{classifier}}(x) \delta x \), an increase in \( \delta x \) directly amplifies \( \Delta L \).
\end{enumerate}

Substituting \( \delta x \) into the expression for \( \Delta L \), we obtain:
\begin{equation}
    \Delta L = \frac{1}{2} \left( J_{\theta} \delta z + \frac{1}{2} \delta z^T H_{\text{decoder}}(z) \delta z + O(\|\delta z\|^3) \right)^T H_{\text{classifier}}(x) \left( J_{\theta} \delta z + \frac{1}{2} \delta z^T H_{\text{decoder}}(z) \delta z + O(\|\delta z\|^3) \right).
\end{equation}

If the second-order term dominates, i.e.,
\begin{equation}
    \delta x \approx \frac{1}{2} \delta z^T H_{\text{decoder}}(z) \delta z,
\end{equation}
then the loss variation simplifies to:
\begin{equation}
    \Delta L = \frac{1}{2} \left( \frac{1}{2} \delta z^T H_{\text{decoder}}(z) \delta z \right)^T H_{\text{classifier}}(x) \left( \frac{1}{2} \delta z^T H_{\text{decoder}}(z) \delta z \right) + O(\|\delta z\|^6).
\end{equation}

For a unit perturbation \( \|\delta z\| = \epsilon \), ignoring higher-order terms, we obtain:
\begin{equation}
    \Delta L \propto \epsilon^4 \cdot \|H_{\text{decoder}}(z)\|^2 \cdot \|H_{\text{classifier}}(x)\|.
\end{equation}

This result confirms that when \( H_{\text{decoder}}(z) \) has large eigenvalues, the classification loss variation \( \Delta L \) increases significantly, making the sample more susceptible to inference attacks. Since \( \|H_{\text{decoder}}(z)\| \) directly scales the extrinsic curvature, samples with higher extrinsic curvature are more vulnerable to MIA. This aligns with the heuristic intuition that high-curvature regions are more sensitive to perturbations, making the model more likely to memorize these samples rather than generalize, which in turn makes it easier for adversaries to exploit membership-related patterns.

\end{document}